\documentclass[10pt,twocolumn,letterpaper]{article}

\usepackage[pagenumbers]{cvpr}

\def\paperID{*****}
\def\confName{CVPR}
\def\confYear{2026}

\usepackage{multirow}
\usepackage{colortbl}
\usepackage{algorithm}
\usepackage{algpseudocode}
\usepackage{wrapfig}
\usepackage{tabularx}
\usepackage{microtype}

\definecolor{cvprblue}{rgb}{0.21,0.49,0.74}
\usepackage[breaklinks,colorlinks,allcolors=cvprblue]{hyperref}

\definecolor{seagreen}{rgb}{0.05,0.52,0.58}
\definecolor{algoblue}{rgb}{0.10,0.30,0.65}
\algrenewcommand\algorithmiccomment[1]{\hfill{\color{seagreen}\itshape\small// #1}}
\algrenewcommand\algorithmicif{\textbf{\color{algoblue}if}}
\algrenewcommand\algorithmicthen{\textbf{\color{algoblue}then}}
\algrenewcommand\algorithmicelse{\textbf{\color{algoblue}else}}
\algrenewcommand\algorithmicfor{\textbf{\color{algoblue}for}}
\algrenewcommand\algorithmicdo{\textbf{\color{algoblue}do}}
\algrenewcommand\algorithmicreturn{\textbf{\color{algoblue}return}}
\algrenewcommand\algorithmicend{\textbf{\color{algoblue}end}}
\algrenewcommand\algorithmicensure{\textbf{\color{algoblue}Ensure:}}
\algrenewcommand\algorithmicrequire{\textbf{\color{algoblue}Require:}}

\begin{document}

\title{ChannelTok: Efficient Flexible-Length Vision Tokenization}

\author{%
  Sukriti Paul \quad Arpit Bansal \quad Tom Goldstein \\
  University of Maryland, College Park \\
  {\tt\small \{sukriti5, bansal01, tomg\}@umd.edu}
}

\maketitle

\begin{abstract}
Leading flexible vision tokenizers achieve SOTA quality at an extreme cost, relying on parameter-heavy backbones and slow, multi-step generative decoders. We depart from this complex, spatial-token paradigm and introduce a simple, lightweight, and fast channel-wise flexible-length tokenizer. Our method treats each latent channel as a visual token, enabling a parameter-efficient CNN-Transformer hybrid backbone. Furthermore, employing a stochastic tail-dropping paradigm during training naturally forces channels to organize by semantic importance.
This allows for flexible compression at inference by simply retaining the first $k$ channels, and naturally enables variable-length autoregressive image generation. We validate our approach through extensive experiments on ImageNet, demonstrating consistent quality across diverse token budgets. The results establish a new quality-efficiency frontier: our model achieves state-of-the-art perceptual quality (rFID 2.92) while being 8.6$\times$ faster in decoding and 2.1$\times$ smaller (159M params) than the next-best alternative. Our work establishes channel-wise tokenization as a powerful and practical paradigm for efficient visual representation. Project page: \url{https://channeltok.github.io}
\end{abstract}
\vspace{-8pt}

\begin{figure}[t]
  \centering
  \vspace{0.8cm}
  \includegraphics[width=\linewidth]{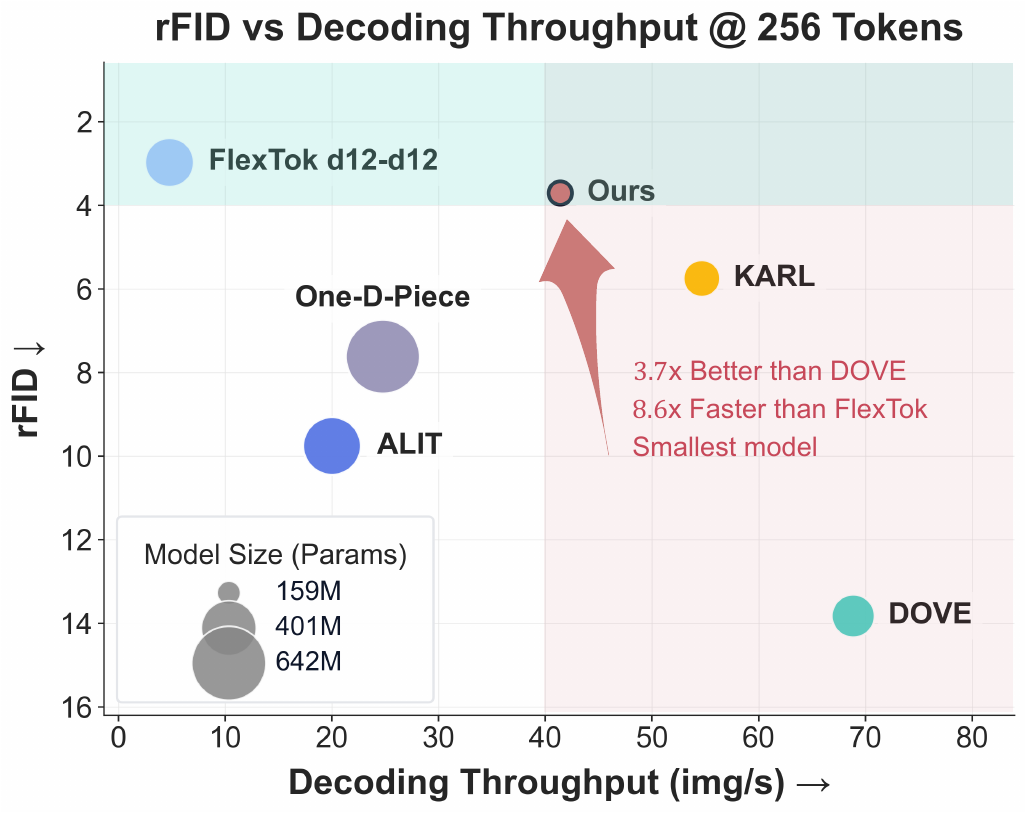}
  \caption{\textbf{Quality-efficiency comparison.} Reconstruction
    fidelity (rFID), decoding throughput, and model size across recent
    tokenizers. Our method achieves state-of-the-art rFID while being
    the smallest and among the fastest decoders.}
    
  \label{fig:teaser}
\end{figure}

\begin{figure*}[t]
\centering
\includegraphics[width=0.9\textwidth]{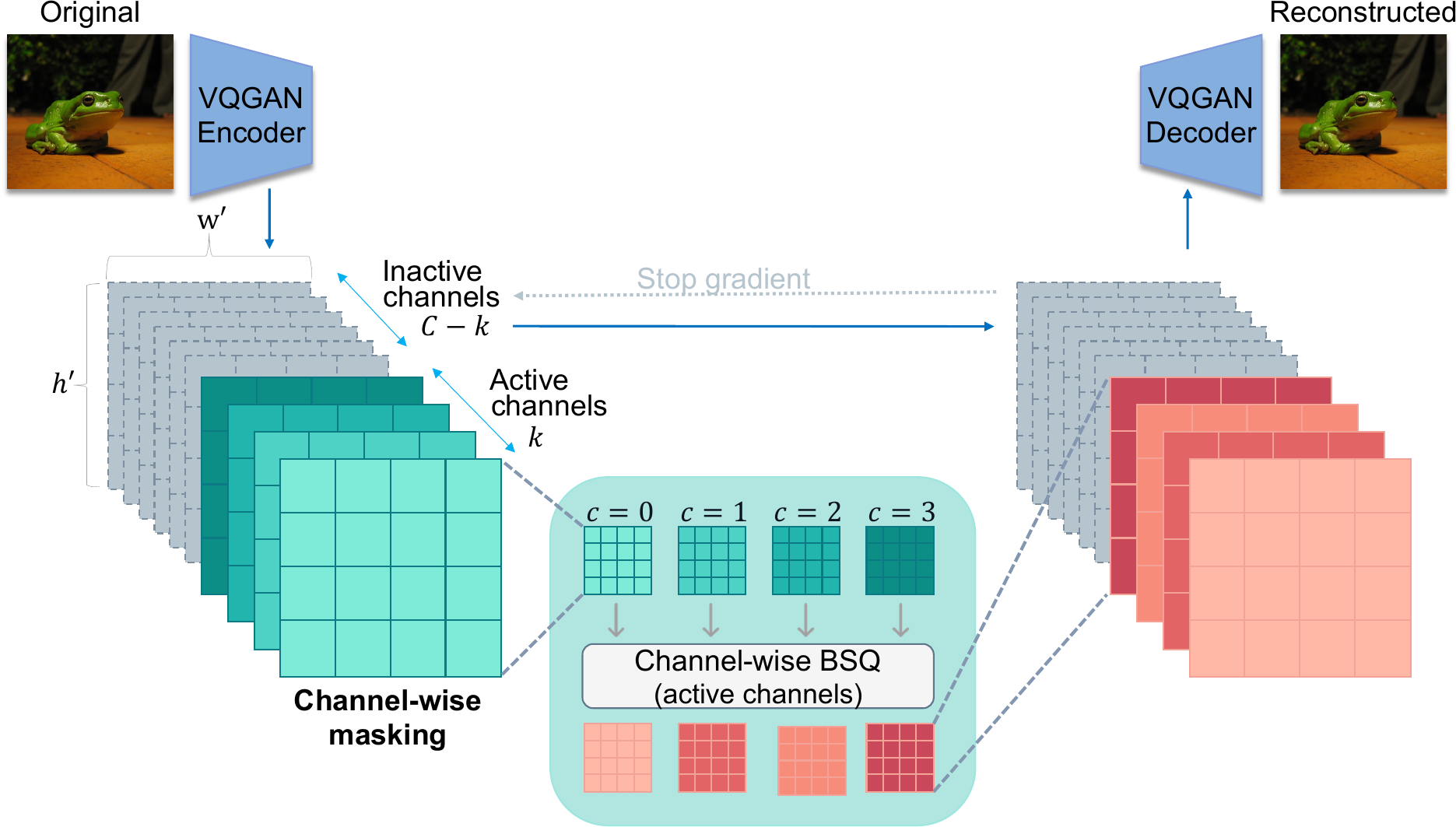}
\caption{\textbf{Overview of our channel-wise flexible tokenizer.} The encoder compresses the input image into a latent representation $\mathbf{z} \in \mathbb{R}^{C \times h \times w}$. During training, we adaptively mask channels by retaining only the first $k$ active channels (shown in teal) while stopping gradients through inactive channels (shown in gray). Each active channel is independently quantized using Binary Spherical Quantization (BSQ). The decoder reconstructs the image from the quantized active channels and the masked inactive channels. At inference, varying $k$ enables flexible compression rates without retraining.}
\label{fig:main_architecture}
\end{figure*}

\section{Introduction}
\label{sec:intro}
Vision tokenization serves as a foundational component in discrete generative image models, compressing images from pixel space into compact, semantically rich representations in latent space. While early generative representations for images focused on continuous-valued autoencoders~\cite{hinton2006reducing} and VAEs~\cite{kingma2014auto}, the pursuit of more efficient and discrete representations led to vector quantization, first via explicit codebooks in VQ-VAE~\cite{oord2017neural} and VQ-GAN~\cite{esser2021taming}, and more recently through lookup-free methods such as FSQ~\cite{mentzer2023finite}, LFQ~\cite{yu2024language}, and BSQ~\cite{zhao2024binary}.

Efficiency in vision tokenization spans multiple dimensions like compression ratio, reconstruction fidelity, semantic richness, encoding-decoding latency, and downstream task performance. Until recently, however, another critical dimension remained largely unexplored: \textit{flexible-length tokenization} that adjusts representation length based on visual complexity. This need has become increasingly urgent in the era of large-scale vision models, where compute budgets constrain deployment and the inherent variability in visual data suggests that not all images require equal representational capacity.

In this work, we introduce a simple {\em channel-wise} approach for training a flexible-length tokenizer. Our approach is based off a lightweight VQGAN~\cite{esser2021taming} based autoencoder where each channel slice forms a visual token. However unlike standard architectures, our latent features are ordered from low- to high-level.  On each step, the encoder maps an image onto the latent space, a random number $k$ of features is chosen, and the decoder must reconstruct the image using only the $k$ lowest-level features.  From this process, a coarse-to-fine hierarchy of features naturally emerges in which ``low-level'' features represent concepts and shapes, and ``high-level'' features represent details.

By leveraging the inherent structure along the channel dimension, our architecture avoids the complexity of parameter-heavy, multi-step generative decoders~\cite{bachmann2025flextok,mao2025dove}. \cref{fig:teaser} shows our method strikes a favorable balance across the quality-efficiency frontier: competitive reconstruction quality (rFID 2.92) while being the smallest (159M parameters) and among the fastest high-fidelity flexible tokenizers. This balance of efficiency and fidelity makes our method ideally suited for deployment scenarios where size, speed, and quality collectively matter, such as edge devices and real-time applications. \\
\newline
\hspace*{1em}Our contributions include: 
(1) a channel-wise flexible tokenization mechanism that dynamically allocates latent channel capacity based on image complexity; (2) a lightweight VQGAN-based architecture optimized for parameter efficiency and high-throughput encoding--decoding; (3) practical training strategies and regularization for stable learning in the channel-dimension tokenization regime; and (4) a demonstration that channel-wise token ordering 
transfers to autoregressive generation, where a LlamaGen model trained 
on our tokens produces coherent images at variable token budgets, 
suggesting channel-wise tokenization as a promising direction for 
efficient AR generation.

\subsection*{Background \& Related Work} 
Recent work has begun to address flexible tokenization through various mechanisms. CAT~\cite{shen2025cat} employs a vision-language model to classify images by complexity level and assigns them to one of three fixed compression ratios within a nested VAE architecture, though this relies on external LLM-based complexity prediction of an image's textual description. Other approaches enforce nested token hierarchies through tail-token dropping~\cite{bachmann2025flextok, miwa2025onedpiece} and block-wise causal masking~\cite{yan2024elastictok}. These Matryoshka-inspired \cite{Kusupati2022MatryoshkaRL} methods demonstrate a semantic hierarchy from coarse to fine representations through explicit token ordering. ALIT~\cite{duggal2024alit} instead uses recurrence to iteratively adjust token lengths over multiple roll-outs, rather than nesting multiple granularities within a single pass. As a further departure, KARL~\cite{duggal2025karl} predicts the appropriate token count in one forward pass guided by Kolmogorov Complexity, framing token length as a proxy for minimal description length. However, it relies on an upside-down reinforcement learning training paradigm and depends heavily on multiple manual design choices for optimizing token budgets, with heightened reconstruction sensitivity to these manual selections.

While these methods expand the notion of ``flexibility'' to encompass reconstruction fidelity, downstream performance, and latent compressibility, industry-scale deployment demands additional considerations: parameter efficiency and inference speed. Many existing approaches rely on sophisticated, parameter-heavy architectures with large ViT backbones~\cite{tian2024detailflow, mao2025dove, bachmann2025flextok}. Some methods, though achieving compelling reconstruction fidelity, require multi-step decoding (FlexTok's~\cite{bachmann2025flextok} rectified
flow decoder, DOVE's~\cite{mao2025dove} transformer-based generative decoder), adding computational
overhead at inference. Others limit compression to fixed token buckets~\cite{duggal2024alit} or train on hundreds of millions of samples~\cite{yan2024elastictok}. Moreover, these approaches typically require extensive training recipes, including multi-stage training pipelines or prolonged training schedules, to maintain quality in low-token regimes.

Contemporary flexible tokenization research for 2D discrete representations primarily operates along the spatial dimension, typically treating local $h \times w$ patches as tokens. Recent fixed-length tokenization work~\cite{zhuang2025wetok, yao2025rectify} has marked a conceptual shift by operating in the channel dimension, revealing that latent channels naturally encode a coarse-to-fine hierarchy that emerges without explicit ordering constraints. Unlike spatial approaches that enforce token ordering through architectural constraints and training objectives, channel-based representations exhibit this hierarchy organically. Motivated by this redefinition of ``visual words,'' we identify the channel dimension as the preferred axis for flexible tokenization, enabling a simpler and more parameter-efficient design.

\begin{table*}[t]
    \centering
    \caption{\textbf{Flexible tokenization results on ImageNet-1K validation set.}
    We report learnable parameters (M = millions), reconstruction metrics (rFID $\downarrow$, LPIPS $\downarrow$, DreamSim $\downarrow$), and system efficiency (encoding/decoding throughput in images/s and latency in ms/image).
    All methods are evaluated at a 256-token budget; we additionally report a 512-token variant of our method.
    Best results are shown in \textbf{bold}, second-best are \underline{underlined}.}
    \label{tab:main_results_256}
    \resizebox{\textwidth}{!}{
    \begin{tabular}{l|ccc|ccc|cccc}
    \toprule
    \multirow{2}{*}{\textbf{Method}}
    & \multicolumn{3}{c|}{\textbf{Parameters (M)}}
    & \multicolumn{3}{c|}{\textbf{Reconstruction Quality}}
    & \multicolumn{4}{c}{\textbf{System Efficiency}} \\
    \cmidrule(lr){2-4} \cmidrule(lr){5-7} \cmidrule(lr){8-11}
    & Encoder & Decoder & Total
    & rFID $\downarrow$ & LPIPS $\downarrow$ & DreamSim $\downarrow$
    & Enc. Tput $\uparrow$ & Dec. Tput $\uparrow$ & Enc. Lat. $\downarrow$ & Dec. Lat. $\downarrow$ \\
    \midrule
    OneDPiece~\cite{miwa2025onedpiece}
    & 304 & 307 & 642
    & 7.61 & 0.180 & \underline{0.109}
    & 32.84 & 24.77 & 30.5 & 40.4 \\

    DOVE~\cite{mao2025dove}
    & \underline{70} & \textbf{70} & 287
    & 13.82 & \textbf{0.153} & 0.133
    & \textbf{118.70} & \textbf{68.83} & \textbf{8.4} & \textbf{14.5} \\

    ALIT~\cite{duggal2024alit}
    & 202 & 206 & 431
    & 9.74 & 0.180 & 0.11
    & 18.51 & 18.34 & 54.01 & 54.54 \\

    KARL~\cite{duggal2025karl}
    & 101 & 105 & \underline{239}
    & 5.74 & \underline{0.154} & 0.116
    & \underline{63.54} & \underline{54.66} & \underline{15.7} & \underline{18.3} \\

    FlexTok~\cite{bachmann2025flextok}
    & 85 & 172 & 341
    & \underline{2.97} & 0.228 & 0.154
    & 61.33 & 4.78 & 16.3 & 209.3 \\

    \rowcolor{gray!15}
    \textbf{Ours (256)}
    & \textbf{65} & \underline{95} & \textbf{159}
    & 3.70 & 0.169 & 0.111
    & 51.34 & 41.39 & 19.5 & 24.2 \\

    \rowcolor{gray!9}
    \textbf{Ours (512)}
    & \textbf{65} & \underline{95} & \textbf{159}
    & \textbf{2.92} & \textbf{0.153} & \textbf{0.096}
    & 51.40 & 41.30 & 19.5 & 24.2 \\

    \bottomrule
    \end{tabular}
    }
\end{table*}

\begin{figure*}[t]
\centering
\begin{subfigure}[b]{0.32\textwidth}
    \centering
    \includegraphics[width=\linewidth]{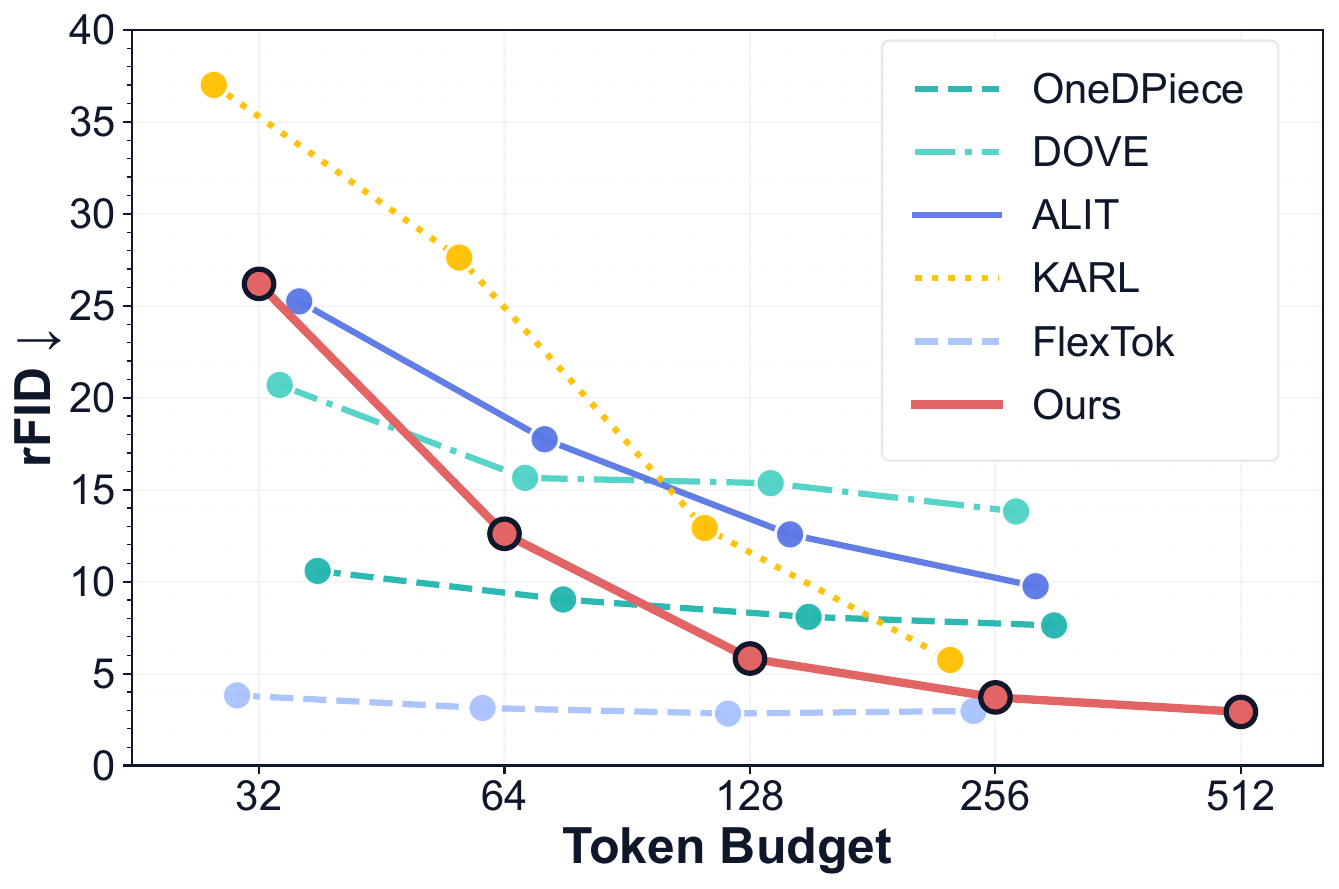}
    \caption{rFID vs.\ Token Budget.}
    \label{fig:pareto_rfid}
\end{subfigure}
\hfill
\begin{subfigure}[b]{0.32\textwidth}
    \centering
    \includegraphics[width=\linewidth]{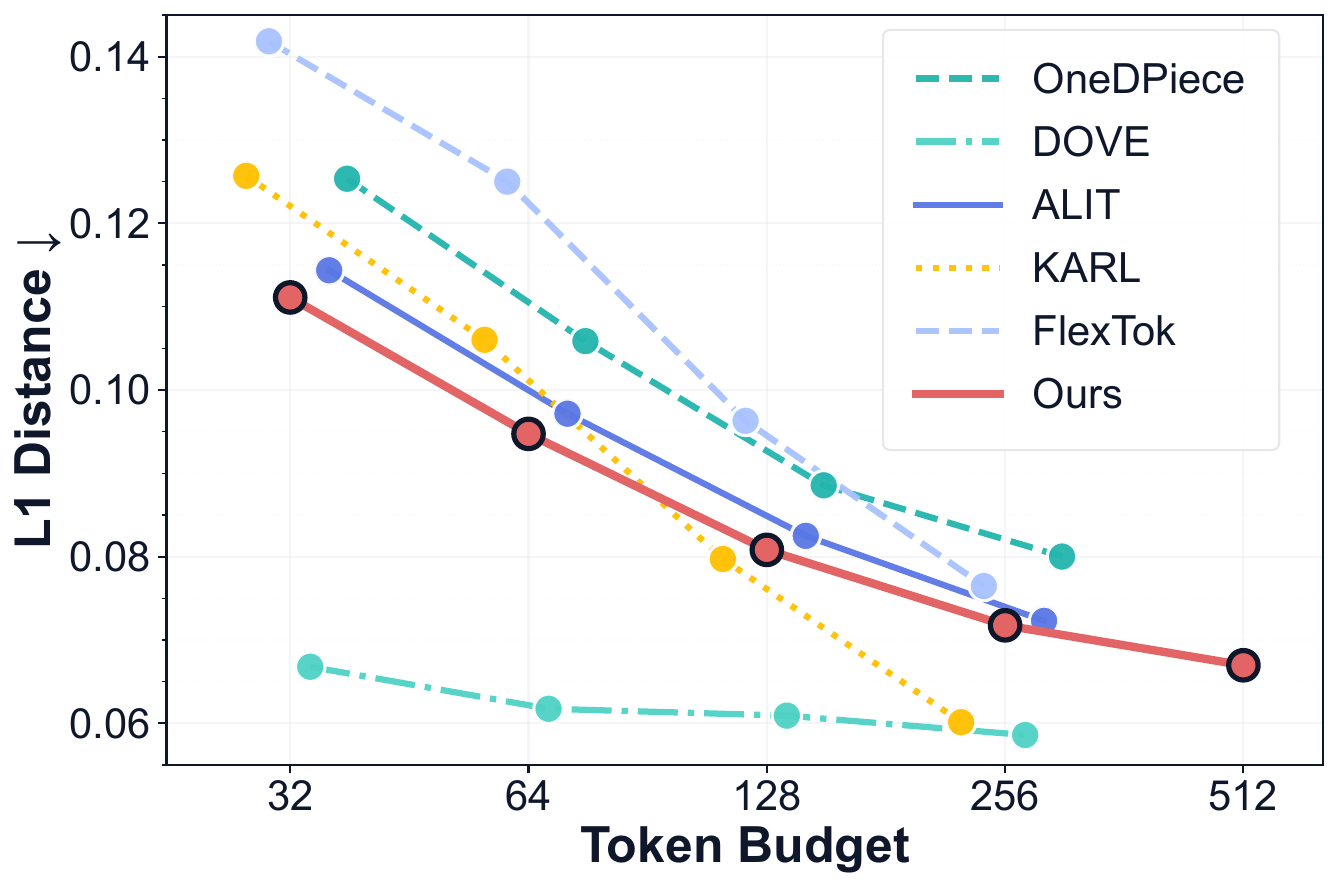}
    \caption{L1 vs.\ Token Budget.}
    \label{fig:pareto_l1}
\end{subfigure}
\hfill
\begin{subfigure}[b]{0.32\textwidth}
    \centering
    \includegraphics[width=\linewidth]{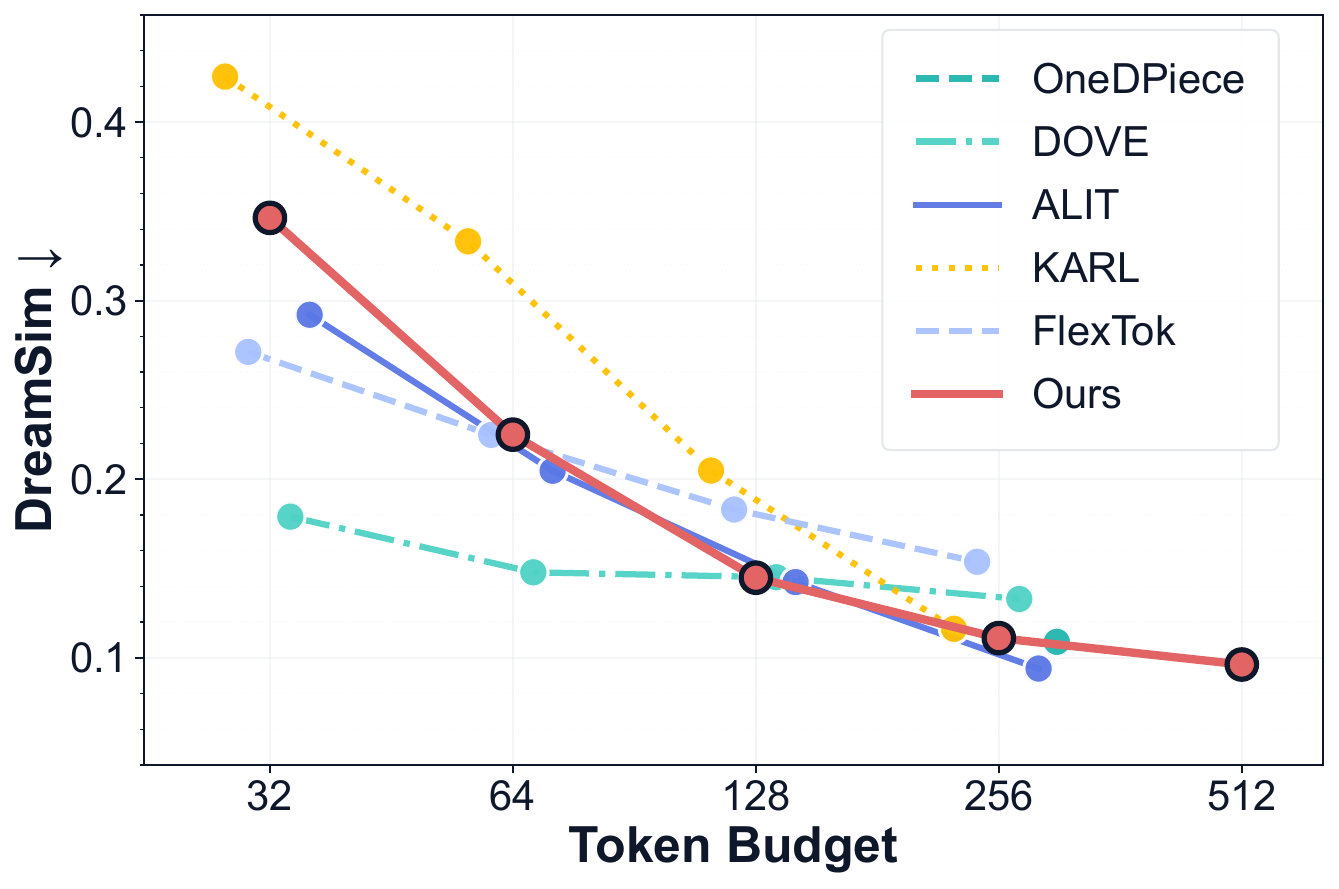}
    \caption{DreamSim vs.\ Token Budget.}
    \label{fig:pareto_dreamsim}
\end{subfigure}
\caption{\textbf{Performance across token budgets.} Our method demonstrates consistent quality improvement across reconstruction metrics across token budgets while being computationally efficient.}
\label{fig:main_quantitative_results}
\end{figure*}

\section{Method}
\label{sec:method}

We introduce a channel-wise adaptive tokenizer that dynamically selects the number of latent channels to retain for each input. Our approach employs a VQGAN-inspired~\cite{esser2021taming} autoencoder where each latent channel functions as a discrete visual token. An overview appears in \cref{fig:main_architecture}.

\subsection{Preliminaries}

Let $\mathbf{x} \in \mathbb{R}^{H \times W \times 3}$ denote an input image. Our tokenizer comprises an encoder $E_\theta$, quantizer $Q_\phi$, and decoder $D_\psi$. The encoder compresses $\mathbf{x}$ into latent representation $\mathbf{z} = E_\theta(\mathbf{x}) \in \mathbb{R}^{C \times h \times w}$, where $C$ is the channel dimension and $h, w$ are spatial dimensions. Unlike spatial tokenization methods treating each spatial location as a token, we operate along the channel dimension where each channel slice $\mathbf{z}_c \in \mathbb{R}^{h \times w}$ constitutes a visual token. This design enables a natural coarse-to-fine hierarchy without explicit ordering constraints.

\subsection{Encoder-Decoder Architecture}

\textbf{Encoder.} Building on VQGAN~\cite{esser2021taming}, our encoder applies convolutional blocks with residual connections and downsampling layers. Multi-head self-attention blocks at resolutions $\{32, 16, 8\}$ capture global dependencies. The encoder reduces spatial resolution by factor 16 (from $256 \times 256$ to $16 \times 16$) while expanding to $C = 512$ channels, yielding $\mathbf{z} \in \mathbb{R}^{512 \times 16 \times 16}$.

\textbf{Channel as token.} Each of the $C$ channels encodes distinct visual information at varying abstraction levels. We treat the channel dimension as the token dimension, where $\mathbf{z}_c$ for $c \in \{1, \ldots, C\}$ represents the $c$-th visual token. This formulation differs fundamentally from spatial tokenization: rather than encoding local patches, each channel captures global patterns across the entire spatial extent, with earlier channels encoding coarse structure and later channels refining details.

\textbf{Decoder.} The decoder mirrors the encoder with upsampling blocks and multi-head self-attention at resolutions $\{8, 16, 32\}$. It reconstructs the image $\hat{\mathbf{x}} = D_\psi(\mathbf{q})$ from quantized representation $\mathbf{q}$ with inactive channels set to zero.

\begin{algorithm}[tb]
\caption{\textbf{Adaptive Channel Masking.}}
\label{alg:channel_mask}
\begin{algorithmic}[1]
\Require Latent $\mathbf{z} \in \mathbb{R}^{C \times h \times w}$, mask probability $p_{\text{mask}}$, range $[t_{\min}, t_{\max}]$
\Ensure Masked latent $\mathbf{z}_{\text{input}}$, mask $\mathbf{M}$, active channels $k$
\State Sample $\text{apply\_mask} \sim \text{Bernoulli}(p_{\text{mask}})$
\If{$\text{apply\_mask} = 1$}
   \State Sample $t \sim \mathcal{U}(t_{\min}, t_{\max})$
   \State $k \gets \max(1, \min(\lfloor t \cdot C \rfloor, C))$
   \State Initialize $\mathbf{M} \in \{0, 1\}^{C \times h \times w}$ with zeros
   \For{$c = 1$ to $k$}
       \State $\mathbf{M}_c \gets \mathbf{1}_{h \times w}$ \Comment{Set first $k$ channels to 1}
   \EndFor
   \State $\mathbf{z}_{\text{active}} \gets \mathbf{M} \odot \mathbf{z}$
   \State $\mathbf{z}_{\text{inactive}} \gets (1 - \mathbf{M}) \odot \text{sg}(\mathbf{z})$ \Comment{Stop-grad on inactive}
   \State $\mathbf{z}_{\text{input}} \gets \mathbf{z}_{\text{active}} + \mathbf{z}_{\text{inactive}}$
\Else
   \State $\mathbf{M} \gets \mathbf{1}_{C \times h \times w}$ \Comment{Use all channels}
   \State $k \gets C$
   \State $\mathbf{z}_{\text{input}} \gets \mathbf{z}$
\EndIf
\State \Return $\mathbf{z}_{\text{input}}, \mathbf{M}, k$
\end{algorithmic}
\end{algorithm}

\subsection{Adaptive Channel Masking}
\label{sec:adaptive_masking}
To enable flexible tokenization, we introduce a stochastic prefix channel masking mechanism during training. This encourages the model to learn an ordered representation where information importance decreases along the channel dimension, following strategies from FlexTok~\cite{bachmann2025flextok}, OneDPiece~\cite{miwa2025onedpiece}, and ElasticTok~\cite{yan2024elastictok}.

\textbf{Mask generation and application.} For each training sample, we uniformly sample mask retention ratio $t \sim \mathcal{U}(t_{\min}, t_{\max})$ with $t_{\min} = 0.002$ and $t_{\max} = 1.0$. The number of active channels is $k = \lfloor t \cdot C \rfloor$, clamped to $[1, C]$. We construct binary mask $\mathbf{M} \in \{0, 1\}^{C \times h \times w}$ as:
\begin{equation}
\mathbf{M}_c = \begin{cases}
1 & \text{if } c \leq k \\
0 & \text{otherwise}
\end{cases}
\end{equation}
The mask will be used to stochastically drop the tail of the feature tensor, promoting hierarchical organization where critical information concentrates in early channels, and fine details in later channels.

 During training, we randomly choose between applying masking or using the full latent $\mathbf{z}$, each with equal probability. For masked channels ($c > k$), we apply stop-gradient to prevent encoding information into inactive channels during the backward pass:
\begin{equation}
\mathbf{z}_{\text{input}} = \mathbf{M} \odot \mathbf{z} + (1 - \mathbf{M}) \odot \text{sg}(\mathbf{z}),
\end{equation}
where $\text{sg}(\cdot)$ denotes stop-gradient. This ensures gradients flow only through active channels, encouraging the encoder to prioritize information in early channels that are more likely to be retained. \cref{alg:channel_mask} formalizes the procedure.

\textbf{Inference flexibility.} At inference, we control compression rate by specifying active channel count $k \in [1, C]$. Given target $k$, we construct mask $\mathbf{M}$ deterministically with first $k$ channels active, encode the image to obtain $\mathbf{z}$, and apply the mask. We quantize only the active channels via BSQ, while inactive channels are set to zero. The decoder reconstructs from both the quantized active channels and the zero-masked inactive channels, enabling continuous control over the rate-distortion tradeoff without retraining. The emergent coarse-to-fine structure allows progressive decoding: reconstructions with $k = 32$ capture global structure, while increasing $k$ progressively refines local details.

\subsection{Binary Spherical Quantization}

We incorporate Binary Spherical Quantization (BSQ)~\cite{zhao2024binary} to discretize each active channel independently. Following the analysis in WeTok~\cite{zhuang2025wetok}, BSQ provides a lookup-free, parameter-efficient alternative to codebook-based methods that is particularly well-suited for channel-wise tokenization. BSQ projects latent features onto a unit hypersphere and applies binary quantization. We use straight-through estimation~\cite{bengio2013estimating} for gradients and optimize entropy-based and commitment losses following~\cite{zhao2024binary}, with $\lambda_{\text{ent}} = 0.1$ and $\lambda_{\text{commit}} = 0.25$ in our quantization loss
\[
\mathcal{L}_{\text{quant}} = \lambda_{\text{ent}}\mathcal{L}_{\text{ent}} + \lambda_{\text{commit}}\mathcal{L}_{\text{commit}}.
\]

\begin{figure*}[t]
\centering
\includegraphics[width=\textwidth]{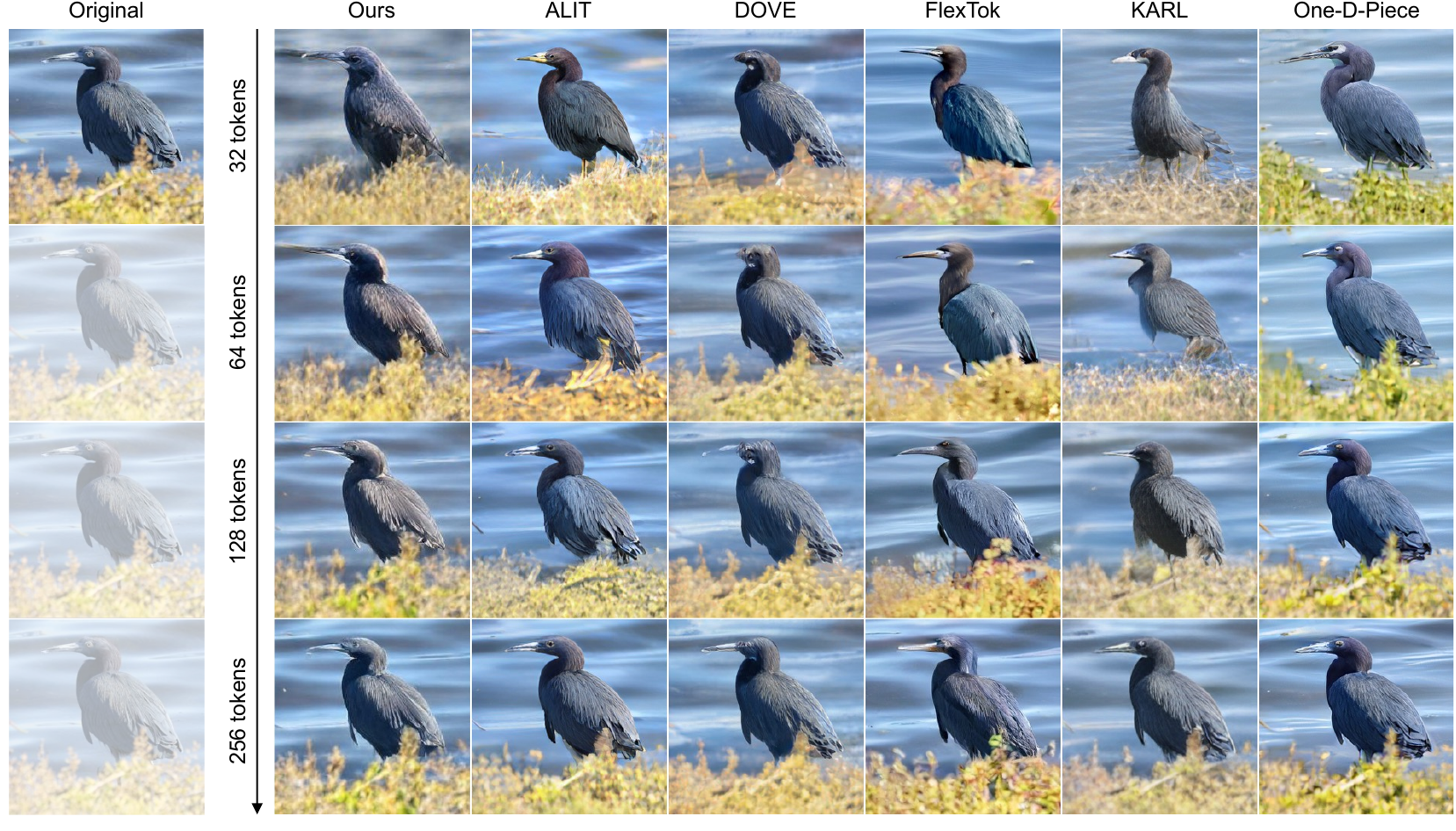}
\caption{\textbf{Qualitative comparison across token budgets.} We show reconstructions from our model and prior flexible tokenizers at 32--256 tokens, along with the original image. Differences in sharpness, color consistency, and structural preservation can be observed as the token budget increases. Additional results are provided in the Appendix.}
\label{fig:qualitative_main}
\end{figure*}

\subsection{Training Objective}

\paragraph{Reconstruction and perceptual losses.}
We supervise the decoder with an $\ell_1$ reconstruction loss $\mathcal{L}_{\text{rec}} = \|\mathbf{x} - \hat{\mathbf{x}}\|_1$ and a VGG-based perceptual loss
\begin{equation}
\mathcal{L}_{\text{perc}} = \sum_{l} \|\Phi_l(\mathbf{x}) - \Phi_l(\hat{\mathbf{x}})\|_2^2,
\end{equation}
where $l$ indexes feature layers of the pretrained network $\Phi$. The base objective combines these with the quantization loss:
\begin{equation}
\mathcal{L}_{\text{base}} = \lambda_{\text{rec}} \mathcal{L}_{\text{rec}} + \lambda_{\text{perc}} \mathcal{L}_{\text{perc}} + \mathcal{L}_{\text{quant}},
\end{equation}
with fixed weights $\lambda_{\text{rec}} = 0.25$ and $\lambda_{\text{perc}} = 1.0$.

\paragraph{Adversarial training.}
To improve perceptual fidelity, we employ a PatchGAN discriminator~\cite{isola2017image} with the standard hinge loss~\cite{miyato2018spectral}. We train for 15 epochs using only $\mathcal{L}_{\text{base}}$ as a warm-up, after which we add the adversarial term and optimize
\[
\mathcal{L} = \mathcal{L}_{\text{base}} + \lambda_{\text{gan}}\mathcal{L}_{\text{hinge}}, \quad \lambda_{\text{gan}}=0.1.
\]
This schedule avoids early adversarial instability while boosting perceptual quality in later stages.

\subsection{Training Considerations}
Flexible-length channel-wise tokenization remains largely unexplored compared to spatial tokenization, with limited established training recipes in the literature. Through experimentation, we identify choices that impact training stability and reconstruction fidelity. We adopt $C = 512$ latent channels to balance capacity with stability. We employ $\ell_1$ reconstruction loss with perceptual loss weighting ($\lambda_{\text{perc}} = 1.0$, $\lambda_{\text{rec}} = 0.25$), where higher perceptual weight enhances reconstruction fidelity; this reversal of traditional weighting aligns with recent findings in visual tokenization that prioritize perceptual quality~\cite{bachmann2025flextok,yu2024language}. We also found $\ell_2$ loss, Charbonnier loss, and gram matrix losses~\cite{wu2024atoken} produced noticeable blurring in our channel-wise setting, while perceptual weights below $0.1$ caused training instabilities.

\begin{figure*}[t]
  \centering
  \includegraphics[width=\textwidth]{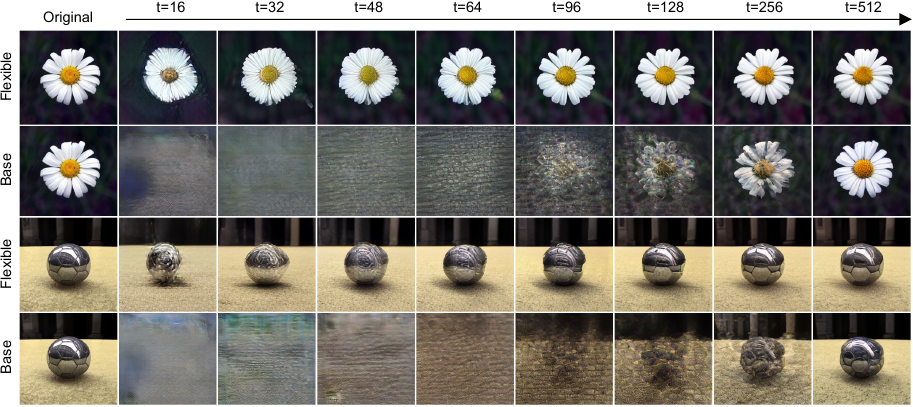}
  \caption{\textbf{Comparison of reconstructions with and without our channel-wise flexible masking module.} The baseline model is identical to the flexible model but trained without masking. For the baseline: semantic coherence appears only at high token budgets, indicating a lack of emergent semantic hierarchy. Our flexible tokenizer clearly shows a semantic coarse-to-fine progression, with meaningful reconstructions even at low and moderate token budgets.}
  \label{fig:progression}
\end{figure*}

We introduce adversarial training to mitigate BSQ quantization artifacts, particularly checkerboard patterns and grid-like distortions that emerge from binary quantization. However, we delay GAN loss introduction until epoch 15, as applying it from initialization led to mode collapse in preliminary experiments. This approach allows the reconstruction objective to establish a stable baseline before adversarial refinement. Our stochastic masking ($p_{\text{mask}} = 0.5$) was more stable than deterministic warmup schemes that maintain full channels for initial epochs before transitioning to adaptive masking. Cosine decay with linear warmup (5K steps to peak LR $10^{-3}$, decaying to $5 \times 10^{-5}$) becomes crucial when combining reconstruction and adversarial objectives~\cite{brock2018large,karras2019style}. Finally, we adopt BSQ with an implicit codebook size of $2^{16}$ over codebook-based vector quantization since our aim was to first establish a training recipe for flexible channel-wise tokenization without additional complexity from codebook commitment and collapse issues~\cite{yu2024language}. BSQ eliminates lookup operations and codebook memory storage, with recent work showing its relevance in channel-wise regimes~\cite{zhuang2025wetok}.

\begin{figure}[t]
  \centering
  \includegraphics[width=\linewidth]{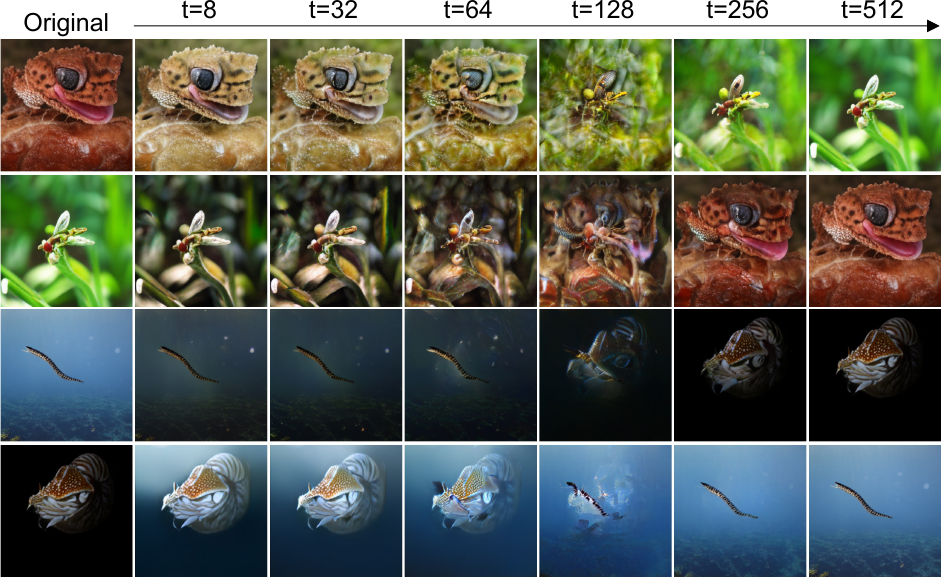}
  \caption{\textbf{Semantic organization in early channels.}
  Channel swapping experiments demonstrate hierarchical semantic encoding. Each pair of rows shows two images whose first $t$ channels are progressively swapped. When channels from one image are replaced with another, the image progressively transforms from the source to the target.}
  \label{fig:semantics}
\end{figure}

\textbf{System Specifications.}
For training, we use 8$\times$ NVIDIA H100 80GB GPUs over a period of 48 hours for training over 150 epochs on ImageNet-1K~\cite{russakovsky2014imagenet}. For our inference, we use 1x NVIDIA A5000 GPU.

\begin{figure*}[t]
\centering
\includegraphics[width=\linewidth]{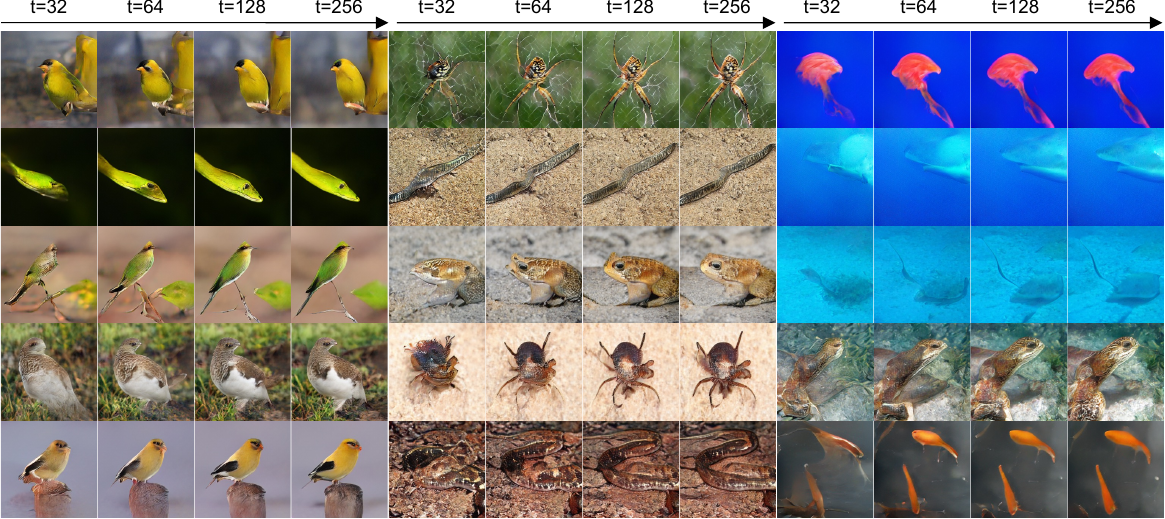}
\caption{\textbf{Autoregressive image generation across token budgets.}
Images are generated by sampling from the LlamaGen~\cite{sun2024autoregressive} GPT-L transformer trained on discrete channel tokens. Generation begins from a randomly sampled first token and proceeds autoregressively, with remaining channels zero-filled at truncation. Even at 32 tokens ($7.9\times$ speedup), generated samples show coherent global structure, with quality improving progressively as more channels are generated.}
\label{fig:generation_images}
\end{figure*}

\section{Experiments}
\label{sec:experiments}

\subsection{Flexible Length Tokenization}

\textbf{Evaluation protocol.} We evaluate reconstruction quality and system efficiency on ImageNet-1K~\cite{russakovsky2014imagenet} validation set (50k images) with standard transforms.
We measure reconstruction fidelity and perceptual quality via PSNR, rFID~\cite{jayasumana2024rfid}, L1 distance, LPIPS~\cite{zhang2018perceptual}, SSIM~\cite{wang2004ssim}, and DreamSim~\cite{fu2023dreamsim}. For system performance, we report encoding/decoding throughput (images/s) and latency (ms/image) on a single A5000 GPU. To ensure a fair and rigorous comparison, all methods were benchmarked with fp16 precision (Note: DOVE required bf16 as it was incompatible with fp16) and a fixed batch size of 128. We isolate model performance to a forward pass by excluding data loading, GPU-CPU transfers, and disk I/O from timing measurements. All baselines were benchmarked under these identical hardware and protocol settings.

\textbf{Performance at 256 tokens.} \cref{tab:main_results_256} demonstrates that our method establishes a new Pareto frontier for flexible tokenization, delivering competitive perceptual quality while being significantly smaller and faster than existing methods. We first evaluate at a 256-token budget to enable fair comparison with key baselines~\cite{bachmann2025flextok,miwa2025onedpiece,duggal2024alit} that support this maximum token count. At 256 tokens, our model achieves an rFID of 3.70, closely matching FlexTok's 2.97. However, our decoder is 8.6$\times$ faster (41.4 vs 4.8 images/s) and our total model is 2.1$\times$ smaller (159M vs 341M params). This efficiency gain stems from our lightweight architecture that avoids the costly multi-step generative decoder used in FlexTok. In contrast, while DOVE achieves high throughput, its perceptual quality (rFID 13.82) is insufficient for high-fidelity reconstruction tasks.
A key advantage of our channel-wise design is the ability to scale capacity without retraining. At our model's full 512-channel capacity (row Ours (512)), we achieve state-of-the-art perceptual quality with an rFID of 2.92, surpassing FlexTok's 2.97. Our method is the only flexible tokenizer that simultaneously achieves best-in-class perceptual quality (rFID 2.92, DreamSim 0.096) and efficiency (smallest model at 159M params, among the fastest decoders), making it ideal for latency-sensitive applications.

\textbf{Performance across token budgets.} \cref{fig:main_quantitative_results} shows reconstruction quality across varying token budgets. Our method demonstrates consistent quality improvement across all metrics, with rFID decreasing from 12.60 at 64 tokens to 2.92 at 512 tokens. At the highest budget, we achieve the best rFID (2.92) and DreamSim (0.096) across all methods, indicating superior semantic preservation and perceptual alignment.
FlexTok maintains consistently low rFID (2.83-3.81) across budgets due to its rectified flow decoder~\cite{bachmann2025flextok}, which provides perceptual coherence largely independent of token count. KARL and ALIT exhibit steeper rFID degradation at lower budgets but achieve strong pixel-level reconstruction at higher token counts, with KARL reaching competitive L1 distance (0.060) at 256 tokens.
The DreamSim metric further validates our approach: scores improve from 0.347 at 32 tokens to 0.096 at 512 tokens, demonstrating that additional channels enable richer feature representations that better capture perceptually-salient content. This trend suggests our channel-wise design naturally scales with capacity, allocating tokens to maximally preserve semantic information. A key advantage of our approach is seamless scaling from 64 to 512 tokens without architectural modifications, enabling flexible quality-efficiency operating points. Visual comparisons across methods and token budgets are provided in \cref{fig:qualitative_main}.

\begin{figure*}[t]
\centering
\begin{subfigure}{0.32\linewidth}
    \includegraphics[width=\linewidth]{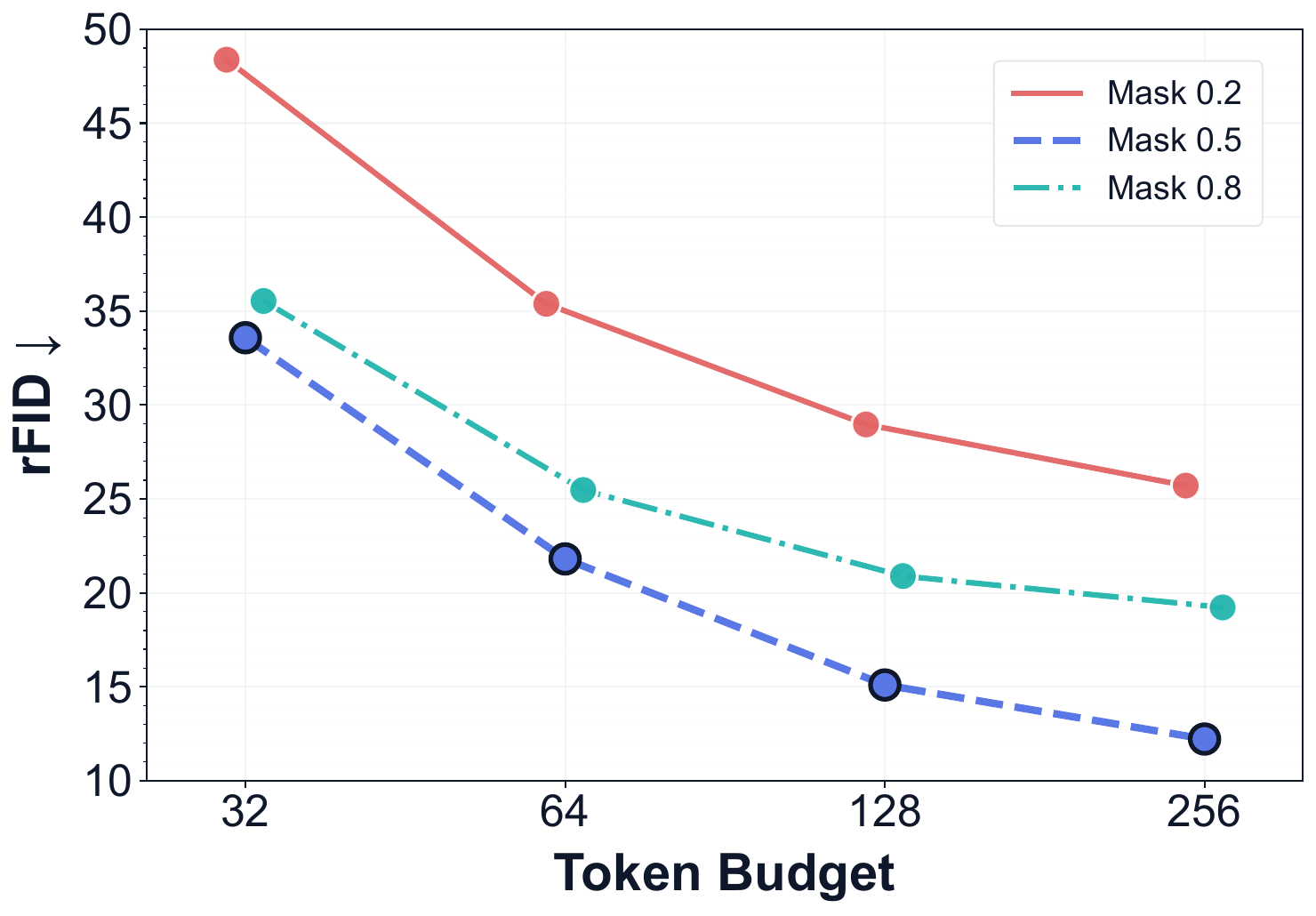}
    \caption{}
    \label{fig:masking_tokens}
\end{subfigure}
\hfill
\begin{subfigure}{0.32\linewidth}
    \includegraphics[width=\linewidth]{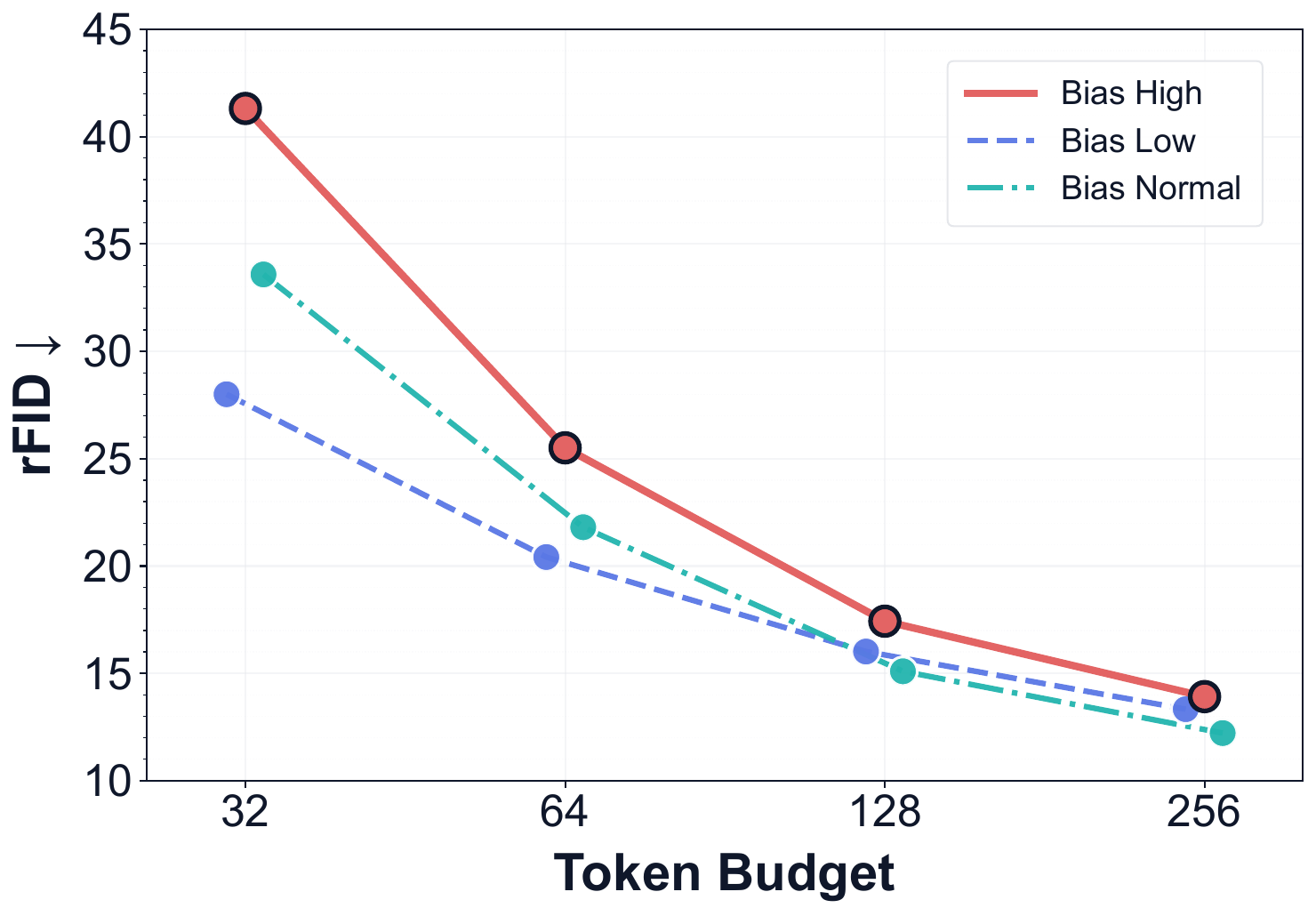}
    \caption{}
    \label{fig:sampling_bias}
\end{subfigure}
\hfill
\begin{subfigure}{0.32\linewidth}
    \includegraphics[width=\linewidth]{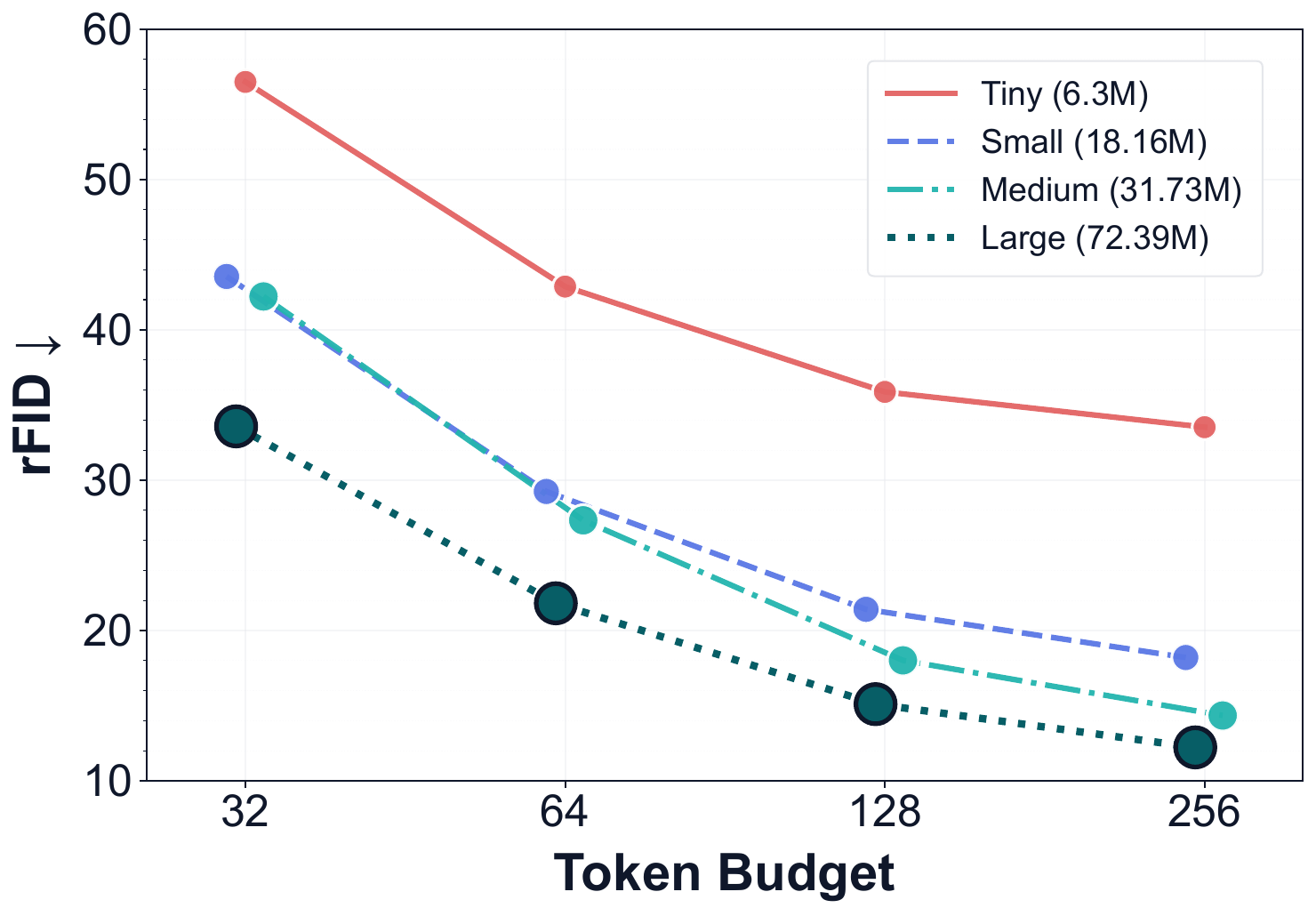}
    \caption{}
    \label{fig:scaling_law}
\end{subfigure}
\caption{\textbf{Architectural ablations.} (a) Effect of masking probability $p_{\text{mask}}$. (b) Effect of sampling bias on retention ratio $t$. (c) Effect of model scale. rFID consistently improves with more balanced masking, uniform sampling, and larger model capacity.}
\label{fig:ablate1}
\end{figure*}

\section{Induced Semantic Hierarchy}
\label{sec:semantics}

\subsection{Prefix Masking Enforces Ordering}
\cref{fig:progression} illustrates the effect of our adaptive masking on the ordering of representations. Our baseline model without tail-dropping produces unstructured channel representations where the progression of tokens lacks sufficient semantic content until the last few tokens. Our tail-dropping strategy forces a causal structure along the channel dimension. Therefore, early channels must encode information independently of later channels, as tail channels may be masked during training. This causal constraint forces our method to concentrate important information in early channels through reconstruction from variable-length prefixes, with later channels adding progressive refinement. Additional training details on our baseline model and qualitative examples are in the supplementary material.



\subsection{Semantic Transferability} We investigate whether the global semantics encoded in early channels transfer between pairs of images. We select visually dissimilar image pairs and swap their first $t$ channels, as shown in \cref{fig:semantics}. For small swaps ($t < 32$), we observe transfer of stylistic elements such as background characteristics, color temperature, and lighting while preserving the primary subject. However, as swap depth increases ($t \geq 64$), images undergo more noticeable semantic transformations with foreground subjects beginning to morph. This behavior mirrors the hierarchical refinement in progressive image coding, where early layers establish global structure before fine details emerge.




\section{Autoregressive Image Generation}
\label{sec:downstream_tasks}           
A key advantage of channel-wise tokenization is its natural compatibility with autoregressive (AR)      
generation. In spatial tokenization, each token encodes a local patch, and the generation order is an
arbitrary raster scan, where truncating the sequence yields an incomplete spatial grid rather than a usable 
image. In contrast, our channel-wise tokens are ordered by semantic importance: each successive token
refines the \emph{entire} image, progressing from coarse structure to fine detail. This ordering maps
directly onto the autoregressive factorization, making variable-length generation not merely possible
but semantically meaningful.

We validate this by first training a tokenizer with a reduced implicit codebook vocabulary ($2^{14}$) and a maximum dimension of 256, for the smaller ImageNet-100 subset. We then train a LlamaGen~\cite{sun2024autoregressive} GPT-L transformer 
on the discrete token sequences extracted by our tokenizer. 
The model is trained with standard next-token prediction using cross-entropy
loss. To reflect the coarse-to-fine hierarchy, we introduce a \textbf{position-weighted} loss that
emphasizes early channels:
\begin{equation}
\mathcal{L}_{\text{AR}} = \sum_{c=0}^{C-1} w_c \cdot \mathcal{L}_{\text{CE}}(c), \quad w_c = 1 +
\alpha\!\left(1 - \frac{c}{C-1}\right),
\label{eq:ar_loss}
\end{equation}
where $\mathcal{L}_{\text{CE}}(c)$ is the cross-entropy at channel position $c$ and $\alpha$ controls
the weighting strength. With $\alpha=1$, weights decrease linearly from $w_0=2.0$ for the first channel
to $w_{C-1}=1.0$ for the last, biasing the model toward accurately predicting the structurally critical
early channels while still learning fine-grained later channels.

  \textbf{Flexible generation budget.} At inference, generation can be terminated after any number of
  tokens $k \leq C$, with remaining channels zero-filled before decoding. \cref{tab:token_budget}
  reports FID (calculated with respect to the tokenizer outputs) and generation throughput across token budgets on a single H100 GPU. At 64 tokens, the model
   achieves FID 9.75 with a $4.1\times$ speedup over full 256-token generation (3.77s $\to$ 0.91s per
  image). The marginal gain from 128 to 256 tokens is negligible (FID 7.96 $\to$ 7.85), consistent with
  the coarse-to-fine hierarchy: early channels carry the dominant semantic content, and later channels
  contribute diminishing perceptual improvement. Even at 32 tokens ($7.9\times$ speedup), generated
  samples exhibit coherent global structure (\cref{fig:generation_images}), confirming that the channel
  ordering learned by the tokenizer transfers effectively to the generative setting.

  This quality-speed tradeoff is architecturally infeasible with spatial tokenizers, where partial
  sequences produce fragmented images. Channel-wise generation uniquely enables a \emph{single} trained
  model to serve diverse latency budgets without retraining or architectural modification.


\begin{table*}[t]
  \centering
  \setlength{\tabcolsep}{6pt}
  \begin{minipage}[t]{0.48\textwidth}
    \vspace{0pt}
    \centering
    \captionof{table}{\textbf{Token budget vs.\ speed/quality} (tokenizer reference).}
    \label{tab:token_budget}
    \footnotesize
    \begin{tabularx}{\linewidth}{@{}l>{\centering\arraybackslash}X>{\centering\arraybackslash}X>{\centering\arraybackslash}X@{}}
      \toprule
      \rule{0pt}{4.9ex} & & & \\[-1.8ex]
      Budget & Speedup & Time/Im. & gFID$^\dagger$ \\
      \midrule
      32  & 7.93$\times$ & 0.475s & 20.71 \\
      64  & 4.13$\times$ & 0.92s  & 9.75  \\
      128 & 2.06$\times$ & 1.82s  & 7.96  \\
      256 & 1.0$\times$  & 3.78s  & 7.85  \\
      \bottomrule
      \multicolumn{4}{l}{\scriptsize $^\dagger$FID w.r.t.\ tokenizer reconstruction.}
    \end{tabularx}
  \end{minipage}\hfill
  \begin{minipage}[t]{0.48\textwidth}
    \vspace{0pt}
    \centering
    \captionof{table}{\textbf{Quantizer performance across token budgets.}}
    \label{tab:quantizer_metrics}
    \footnotesize
    \begin{tabularx}{\linewidth}{@{}l>{\centering\arraybackslash}X>{\centering\arraybackslash}X>{\centering\arraybackslash}X>{\centering\arraybackslash}X@{}}
      \toprule
      & \multicolumn{2}{c}{FSQ} & \multicolumn{2}{c}{BSQ} \\
      \cmidrule(lr){2-3} \cmidrule(l){4-5}
      Budget & PSNR & L1 & PSNR & L1 \\
      \midrule
      32  & 15.37 & 0.1166 & 15.58 & 0.1158 \\
      64  & 16.33 & 0.1020 & 16.42 & 0.1031 \\
      128 & 17.46 & 0.0881 & 17.34 & 0.0911 \\
      256 & 18.42 & 0.0775 & 18.04 & 0.0827 \\
      \bottomrule
      \multicolumn{5}{l}{\scriptsize PSNR in dB; L1 is mean absolute error.}
    \end{tabularx}
  \end{minipage}
\end{table*}

\section{Ablations}
\label{sec:ablations}

\subsection{Effects of Sampling and Masking}
\label{subsec:sampling}
As described in \cref{sec:adaptive_masking}, our architecture introduces two forms of stochasticity: a global masking probability $p_{\text{mask}}$ and a per-sample retention ratio $t$. We sweep $p_{\text{mask}}$ on ImageNet-100 and measure rFID across token budgets. Low masking ($p_{\text{mask}} = 0.2$) performs worst as the model overfits to full-context reconstructions, while high masking ($p_{\text{mask}} = 0.8$) improves robustness but slightly hurts fidelity. We find $p_{\text{mask}} = 0.5$ optimal, providing sufficient masking pressure to induce semantic structure without destabilizing reconstructions.
We then vary the retention distribution $t$ at fixed $p_{\text{mask}} = 0.5$ (\cref{fig:sampling_bias}). Biasing $t$ towards lower or higher values trades off performance between token regimes. Uniform sampling provides the most balanced training signal across all budgets.

\subsection{Scaling Parameters}
\label{subsec:scaling}
We train four models by varying the parameters of both the encoder and decoder. The trained models clearly exhibit a scaling curve, with larger models consistently improving rFID scores across token budgets.

\subsection{Effect of Quantizers}
\label{subsec: quantizers}
To study the effect of different quantizers we also train our tokenizer with FSQ on ImageNet-100~\cite{mentzer2023finite} and compare it with BSQ in \cref{tab:quantizer_metrics}. We observe that the channelwise tokenization is agnostic to the quantization method.




\section{Conclusion}
\label{conclusion}
We introduce a channel-wise flexible tokenizer that achieves excellent rFID with high image throughput, offering the best trade-off between quality and efficiency. Our key insight is that coarse-to-fine semantic hierarchy emerges naturally when tokenizing in the channel dimension, eliminating the need for complex architectural constraints. Future work includes exploring adaptive channel selection mechanisms for task-specific optimization and investigating training strategies to further improve low-token regime performance.

\section*{Acknowledgments}

This work was supported by DOE Office of Science’s ASCR AI for Science initiative, the NSF TRAILS Institute (2229885), and Coefficient Giving, and Longview Philanthropy.

{\small
\bibliographystyle{ieeenat_fullname}
\bibliography{main}
}

\clearpage
\appendix
\maketitlesupplementary


\section{Training and Evaluation Details}
\subsection{Training}
We provide training specifications to facilitate reproducibility in
\cref{tab:architecture}, \cref{tab:training}, and \cref{tab:losses}. All training experiments for flexible and baseline tokenizer models were conducted on 8$\times$ NVIDIA H100 80GB GPUs using PyTorch with distributed data parallel (DDP) training. Our baseline model is architecturally identical to our flexible tokenizer model, but devoid of the flexible module (causal masking).

\subsection{Training Stability Notes}
Several design choices were critical for stable training:

\begin{itemize}
    \item \textbf{Two-stage GAN introduction:} Applying adversarial loss from initialization caused mode collapse. Delaying until epoch 15 allows reconstruction objectives to establish a stable baseline.
    \item \textbf{Stochastic masking:} Deterministic warmup schemes (using full channels for initial epochs) were less stable than our stochastic approach with $p_{\text{mask}} = 0.5$.
    \item \textbf{Learning rate schedule:} Cosine decay with warmup is essential when combining reconstruction and adversarial objectives.
    \item \textbf{Gradient clipping:} Max norm clipping at 1.0 prevents gradient explosions during masked training.
    \item \textbf{Loss weighting:} Higher perceptual weight ($\lambda_{\text{perc}} > \lambda_{\text{rec}}$) improves rFID; values below 0.1 cause instabilities.
\end{itemize}

\subsection{Evaluation}
We evaluate reconstruction quality and system efficiency on the ImageNet-1K validation set (50,000 images). All images undergo standard preprocessing: resizing to $256 \times 256$ and normalization with mean and standard deviation of [0.5, 0.5, 0.5] across RGB channels. We save preprocessed \textit{original} images as lossless PNG files to ensure consistent rFID computation.

Reconstruction fidelity is assessed through complementary metrics that are computed using established libraries: LPIPS with VGG backbone via lpips package\footnote{\scriptsize\raggedright\url{https://github.com/richzhang/PerceptualSimilarity}}, SSIM via torchmetrics\footnote{\scriptsize\raggedright\url{https://lightning.ai/docs/torchmetrics/}}, and rFID via clean-fid\footnote{\scriptsize\raggedright\url{https://github.com/GaParmar/clean-fid}}. For rFID computation, both original and reconstructed images are saved as lossless PNG files following the preprocessing protocol above.

\begin{figure}[t]
  \centering
  \includegraphics[width=\linewidth]{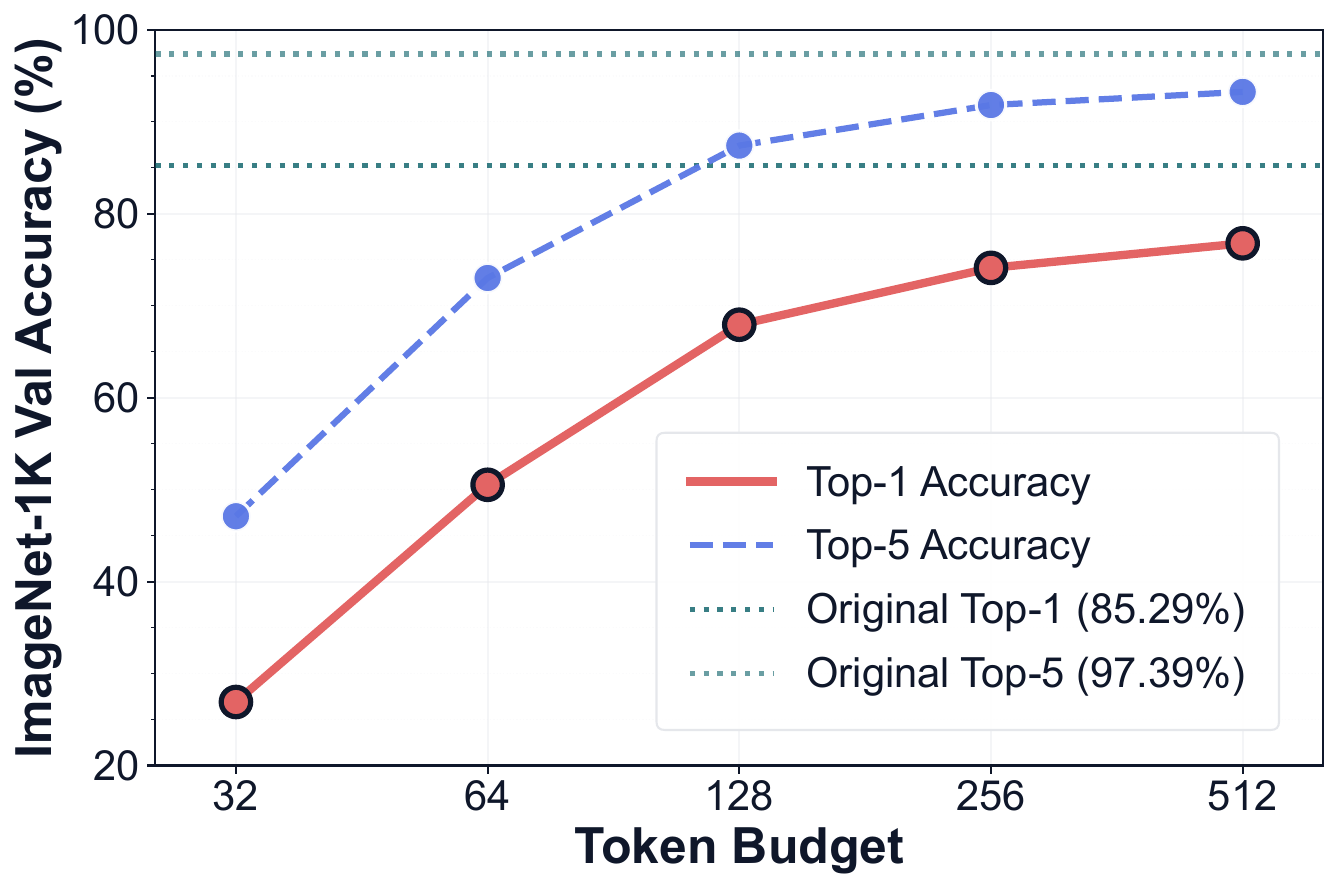}
  \caption{\textbf{DINOv2 classification accuracy across token budgets.} Higher token budgets preserve more discriminative structure, mirroring the rFID trends.}
  \label{fig:dinov2}
\end{figure}

\begin{table*}[t]
\centering
\caption{\textbf{Model architecture specifications.}}
\label{tab:architecture}
\footnotesize
\setlength{\tabcolsep}{4pt}
\resizebox{\linewidth}{!}{
\begin{tabular}{ll|ll|ll}
\toprule
\multicolumn{2}{c|}{\textbf{Encoder (65M)}} & \multicolumn{2}{c|}{\textbf{Decoder (95M)}} & \multicolumn{2}{c}{\textbf{BSQ Quantizer}} \\
\midrule
Input resolution & $256^2 \times 3$ & Input resolution & $16^2 \times 512$ & Method & Binary Spherical \\
Latent resolution & $16^2 \times 512$ & Output resolution & $256^2 \times 3$ & Codebook size & $2^{16}$ \\
Downsampling & $16\times$ & Upsampling & $16\times$ & Dimension & $16$ ($4 \times 4$) \\
Channels & [192,256,384,512,640] & Channels & [640,512,384,256,192] & Channel-wise & True \\
Attention res. & \{32,16,8\} & Attention res. & \{8,16,32\} & Lookup-free & Yes \\
Attention heads & 8 (Flash) & Attention heads & 8 (Flash) & & \\
Residual blocks & 2 per res. & Residual blocks & 3 per res. & \multicolumn{2}{c}{\textbf{Total: 159M params}} \\
\bottomrule
\end{tabular}
}
\end{table*}

\begin{table*}[t]
\centering
\caption{\textbf{Training configuration and hyperparameters.}}
\label{tab:training}
\footnotesize
\setlength{\tabcolsep}{4pt}
\resizebox{\linewidth}{!}{
\begin{tabular}{ll|ll|ll}
\toprule
\multicolumn{2}{c|}{\textbf{Optimization}} & \multicolumn{2}{c|}{\textbf{Training Schedule}} & \multicolumn{2}{c}{\textbf{Data Pipeline}} \\
\midrule
Optimizer & AdamW (fused) & Total epochs & 150 & Dataset & ImageNet-1K \\
Peak LR & $10^{-3}$ & Batch/GPU & 32 & Train images & 1.28M \\
Discriminator LR & $10^{-4}$ & Effective batch & 256 & Val images & 50K \\
Weight decay & 0.01 & Num workers & 8 & Data format & WebDataset \\
Gradient clip & 1.0 (max norm) & Precision & bfloat16 mixed & Loader & DALI (GPU) \\
LR schedule & Cosine + warmup & DDP & True & Train aug. & Crop + flip \\
Warmup steps & 5,000 & Sync BN & True & Val aug. & Resize + crop \\
Init LR & $10^{-6}$ & Compile & torch.compile & Normalize & [0.5,0.5,0.5] \\
Min LR & $5 \times 10^{-5}$ & Flash Attn & Enabled & & \\
\midrule
\multicolumn{6}{c}{\textbf{System: 8$\times$ NVIDIA H100 80GB, 48 hours, PyTorch 2.1+}} \\
\bottomrule
\end{tabular}
}
\end{table*}

\begin{table*}[t]
\centering
\caption{\textbf{Loss configuration and channel masking.}}
\label{tab:losses}
\footnotesize
\setlength{\tabcolsep}{4pt}
\resizebox{\linewidth}{!}{
\begin{tabular}{lll|ll}
\toprule
\multicolumn{3}{c|}{\textbf{Loss Components}} & \multicolumn{2}{c}{\textbf{Channel Masking}} \\
\textbf{Component} & \textbf{Weight} & \textbf{Schedule} & \textbf{Parameter} & \textbf{Value} \\
\midrule
\multicolumn{3}{c|}{\textit{Stage 1: Reconstruction (Epochs 0-14)}} & Mask prob. $p_{\text{mask}}$ & 0.5 \\
L1 reconstruction & $\lambda_{\text{rec}} = 0.25$ & All epochs & Min retention $t_{\min}$ & 0.002 \\
Perceptual (LPIPS) & $\lambda_{\text{perc}} = 1.0$ & All epochs & Max retention $t_{\max}$ & 1.0 \\
BSQ entropy & $\lambda_{\text{ent}} = 0.1$ & All epochs & Sampling & $\mathcal{U}(t_{\min}, t_{\max})$ \\
BSQ commitment & $\lambda_{\text{commit}} = 0.25$ & All epochs & Total channels $C$ & 512 \\
BSQ diversity $\gamma$ & 1.0 & All epochs & Active channels $k$ & $\max(1, \lfloor t \cdot C \rfloor)$ \\
\multicolumn{3}{c|}{\textit{Stage 2: + Adversarial (Epoch 15+)}} & Stop-gradient & True (inactive) \\
Generator adversarial & $\lambda_{\text{gan}} = 0.1$ & Epoch $\geq$ 15 & & \\
Discriminator & PatchGAN (3 layers, 64 ch) & Epoch $\geq$ 15 & & \\
Adversarial loss & Hinge loss & Epoch $\geq$ 15 & & \\
\bottomrule
\end{tabular}
}
\end{table*}

\begin{table*}[t]
\centering
\caption{\textbf{Autoregressive Generation configuration and hyperparameters.}}
\label{tab:ar_training}
\footnotesize
\setlength{\tabcolsep}{4pt}
\resizebox{\linewidth}{!}{
\begin{tabular}{ll|ll|ll}
\toprule
\multicolumn{2}{c|}{\textbf{Architecture (GPT-L)}} & \multicolumn{2}{c|}{\textbf{Optimization}} & \multicolumn{2}{c}{\textbf{Flexible Generation Setup}} \\
\midrule
Parameters & 343M & Optimizer & AdamW (fused) & Training length & 512 (Full Budget) \\
Layers & 12 & Peak LR & $8 \times 10^{-4}$ & Inference length & Dynamic (32--512) \\
Attention heads & 12 & Weight decay & 0.05 & Objective & Weighted Cross-Entropy \\
Hidden dimension & 768 & Betas ($\beta_1, \beta_2$) & (0.9, 0.95) & Early token reward & High \\
Vocab size & 16384 & Gradient clip & 1.0 (max norm) & Resid/FFN Dropout & 0.1 \\
Max seq. length & 512 & LR schedule & Cosine & Class drop prob. & 0.1 \\
Pos. Encoding & 1D Absolute/RoPE & Warmup fraction & 0.1 (10\%) & Token dropout & 0.1 \\
Hardware & 8$\times$ H100 & & & & \\
\bottomrule
\end{tabular}
}
\end{table*}

\section{Downstream Analysis}
\label{sec:downstream_supplementary}

\subsection{Autoregressive Image Generation (LlamaGen)}

To evaluate the efficacy of our flexible-length visual tokens, we train an autoregressive (AR) generation model following the LlamaGen framework. We adopt a GPT-L (Large) architecture with 343M parameters. The model is trained unconditionally or class-conditionally on pre-extracted latents of ImageNet-100.

\paragraph{Flexible-Length AR Training Objective.}
Unlike standard fixed-length tokenization which treats all spatial tokens equally, our autoregressive training utilizes the full token sequence (e.g., 512 tokens) but applies a temporally weighted Cross-Entropy loss. This weighting scheme assigns higher importance to earlier predictions in the sequence, forcing the AR model to prioritize generating the most critical structural and semantic information first. Because the network is explicitly rewarded for early accuracy, we can simply truncate (or "chop") the generation process at inference time at any desired token budget ($t \leq t_{\max}$) to achieve the desired compute-quality trade-off, entirely avoiding the need for variable-length padding or masking during training.

\paragraph{Optimization Specifications.}
We optimize the model using fused AdamW with a peak learning rate of $8 \times 10^{-4}$. We utilize a cosine decay learning rate schedule with a 10\% linear warmup. For class-conditional models, classifier-free guidance is enabled by dropping the class label with a 10\% probability. Detailed hyperparameters are provided in \cref{tab:ar_training}.



\begin{figure*}[t]
  \centering
  \includegraphics[width=\linewidth]{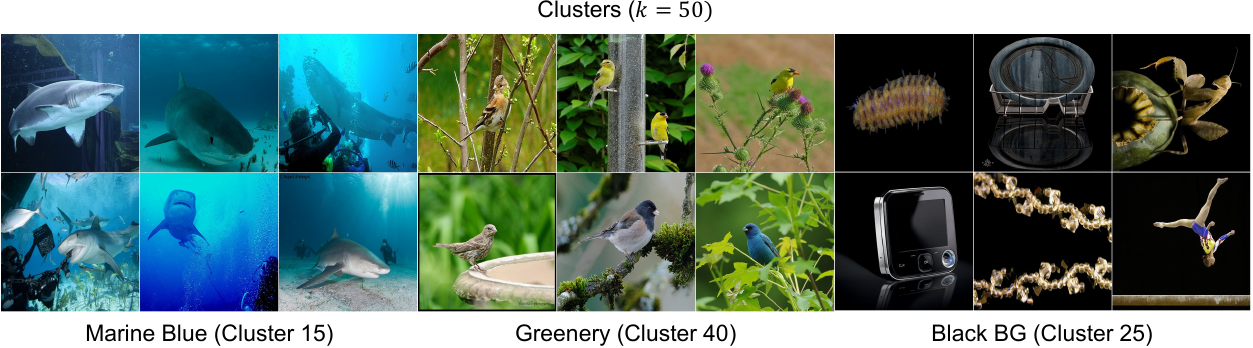}
  \caption{\textbf{Semantic clustering of early channels.} K-means clustering on first 32 channels produces semantically coherent groups organized by scene characteristics (black, marine blue, greenery), hinting that early channels encode meaningful semantic structure. Beyond global scene attributes, clusters also align with object-level semantics, grouping marine life and birds into distinct regions. Crucially, this organisation is never explicitly supervised: it emerges purely from the channel-wise masking objective, suggesting that prioritising early channels during training naturally induces a semantically ordered latent space.}
  \label{fig:cluster}
\end{figure*}
\newpage
\subsection{Semantic Feature Preservation}
To measure how well our tokenizer retains semantic information, we evaluate reconstructed images using a pretrained DinoV2~\cite{oquab2023dinov2} backbone on the ImageNet-1K validation set. As shown in~\cref{fig:dinov2}, both Top-1 and Top-5 accuracy increase steadily with the reconstruction token budget, indicating that higher budgets preserve more discriminative structure. This trend closely follows the rFID improvements reported earlier, confirming that perceptual fidelity and semantic retention improve hand-in-hand in our model.

\subsection{Semantic Organization in Early Channels}
We perform $K$-means clustering ($K=50$) on the first 32 channels of quantized latents from the ImageNet-1K validation set. As shown in \cref{fig:cluster}, clusters exhibit coherent semantic groupings along two axes: (1) global scene characteristics (marine blues, verdant backgrounds, warm palettes, high-contrast compositions) and (2) object-level semantics (aquatic life, reptiles, birds, canines, produce). These groupings emerge organically without class supervision, confirming early channels encode coarse categorical structure. A Calinski-Harabasz~\cite{calinski1974dendrite} score of 404.7 confirms well-separated centroids across ImageNet's categories.


\section{Per-Class Token Budget Analysis}
We conduct an analysis of token requirements across all 1,000 ImageNet-1K validation classes to understand our tokenizer's flexible allocation behavior across different classes. For each class, we determine the minimum number of tokens required to achieve a perceptual quality threshold (LPIPS$\leq$ 0.15). Aggregating statistics yields per-class distributions characterized by mean, revealing the relationship between semantic content and representational capacity.

\begin{figure*}[tb]
\centering
\includegraphics[width=\linewidth]{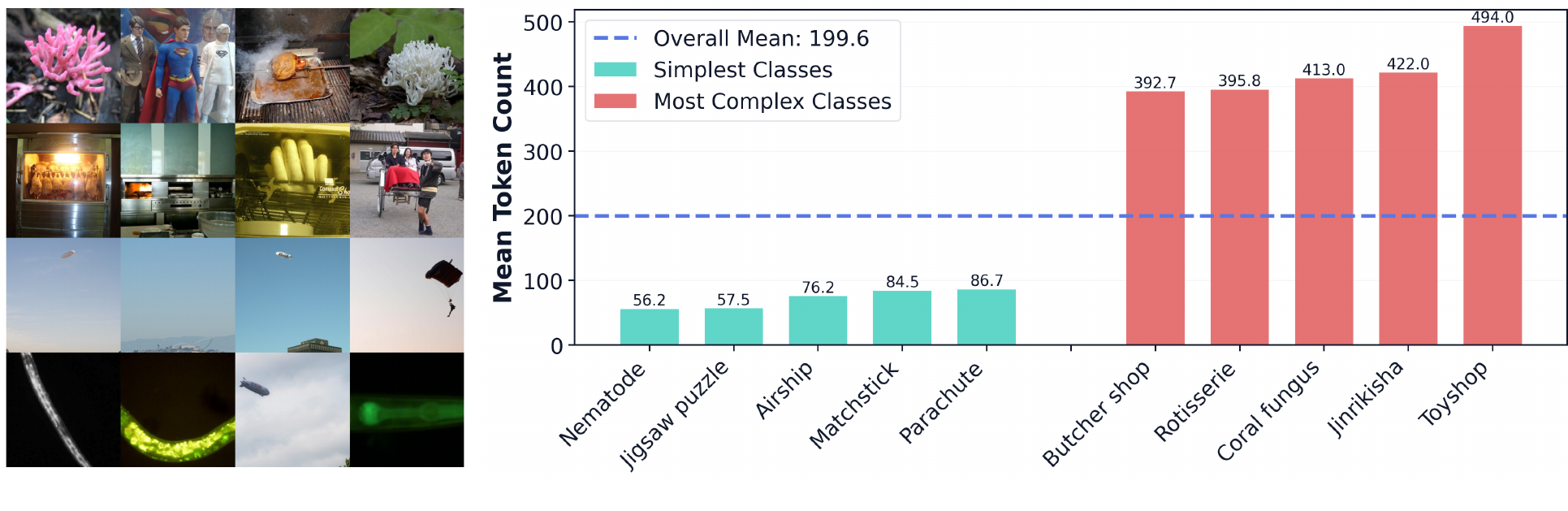}
\caption{\textbf{Token allocation across ImageNet-1K validation classes.}
\textit{Left:} Rows 1--2 show complex classes that require high token counts: Coral fungus (498, 490), Toyshop (494), Rotisserie (491, 459, 442, 432), and Jinrikisha (422), all featuring intricate textures and fine-grained details. Rows 3--4 show visually simple classes that need far fewer tokens: Airship (5, 6, 9, 12), Parachute (9), and Nematode (12, 12, 13), characterised by uniform backgrounds and simple structures.
\textit{Right:} Per-class mean token counts for the five simplest and five most complex classes. Teal bars denote simple classes, red bars complex classes, and the dashed line marks the dataset mean (199.6 tokens). The clear separation between the two groups shows that our tokenizer adapts its budget to visual complexity rather than allocating tokens uniformly across categories. The complexity ordering also aligns with human intuition, as classes that people would judge as visually intricate (dense textures, cluttered scenes) consistently demand more tokens, while perceptually simple classes (plain backgrounds, minimal structure) require far fewer.}
\label{fig:class_analysis}
\end{figure*}

\cref{fig:class_analysis} shows the correlation between visual complexity and token allocation. Geometrically simple classes such as Airship, Parachute, and Nematode consistently achieve the perceptual threshold with mean token budgets of only $56$--$87$ tokens, well below the global per-class mean of $199.6$ tokens. Conversely, classes with intricate textures and fine-grained details, including Coral fungus, Toyshop, Rotisserie, and Jinrikisha, require considerably higher budgets of $392$--$494$ tokens on average. \cref{fig:class_analysis} summarizes this gap by plotting the mean token counts for the five simplest and five most complex classes, making the inter-class stratification explicit: token budgets align with visual complexity rather than being uniformly allocated across categories.

\begin{figure*}[tb]
\centering
\includegraphics[width=\linewidth]{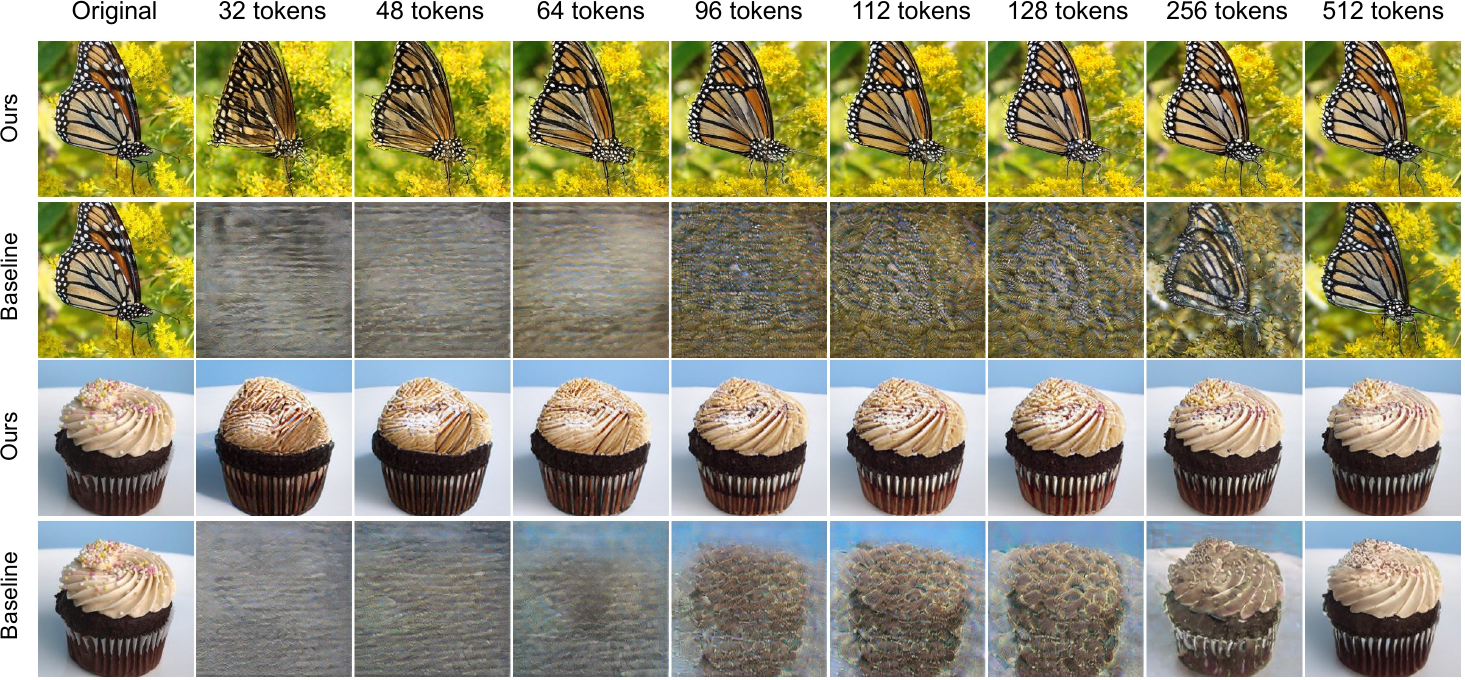}
\caption{\textbf{Reconstruction with and without prefix masking.} Each image pair shows channel progression across increasing token budgets. The first row is our flexible tokenizer and the second row is the baseline, which is architecturally identical but trained without channel-wise adaptive masking. Without masking, the baseline produces no meaningful reconstruction at low token counts, with recognisable structure emerging only at high budgets. Our flexible tokenizer, by contrast, exhibits a clear coarse-to-fine progression, recovering global semantics early and refining detail as more channels are added.}
\label{fig:channel_prog_supp}
\end{figure*}

\section{Ablations}
We conduct ablations (\cref{sec:ablations}) on ImageNet-100 to analyze key design choices. For scaling trends (\cref{subsec:scaling}), we train four model sizes: Tiny (6.3M), Small (18.16M), Medium (31.73M), and Large (72.39M). We perform network-width ablation by progressively increasing channel dimensions of convolutional layers in the encoder-decoder backbone, while keeping latent space (256-D), quantizer (BSQ with 65,536 codes), training schedule, and loss configuration fixed.

\cref{subsec:sampling} investigates sampling bias effects on retention ratio $t$ at fixed $p_{\text{mask}}=0.5$ using three piecewise-uniform distributions:
\begin{itemize}[topsep=2pt, itemsep=1pt, leftmargin=*]
\item \texttt{bias\_lower}: Samples from lower 25\% of $[t_{\min}, t_{\max}]$ with probability 0.75, remainder otherwise
\item \texttt{bias\_higher}: Symmetrically emphasizes upper 25\% of $[t_{\min}, t_{\max}]$ with probability 0.75
\item \texttt{uniform}: Standard uniform distribution on $[t_{\min}, t_{\max}]$
\end{itemize}

To isolate the gains of the channel-wise paradigm from the quantizer choice, we run our tokenizer with FSQ~\cite{mentzer2023finite} as a drop-in replacement for BSQ (\cref{subsec: quantizers}). Both yield similar trends across budgets, confirming the improvements stem from the channel-wise paradigm, not the quantizer.

\begin{table*}[tb]
    \centering
    \caption{\textbf{Reconstruction quality across token budgets on the ImageNet-1K validation set.}
    We report LPIPS $\downarrow$, SSIM $\uparrow$, and PSNR $\uparrow$ at five token budgets.
    Our method is the only one that scales to 512 tokens, where it achieves LPIPS 0.153 (matching the best score of any competing method) with a 1.8$\times$ lighter model (159M vs.\ 287M parameters).
    Quality improves monotonically at every budget (LPIPS: $0.344\!\to\!0.153$), reflecting the coarse-to-fine channel hierarchy learned during training.}
    \label{tab:token_budget_results_1}

    \small
    \resizebox{\linewidth}{!}{
    \begin{tabular}{l|ccc|ccc|ccc|ccc|ccc}
    \toprule
    \multirow{2}{*}{\textbf{Method}}
    & \multicolumn{3}{c|}{\textbf{32 Tokens}}
    & \multicolumn{3}{c|}{\textbf{64 Tokens}}
    & \multicolumn{3}{c|}{\textbf{128 Tokens}}
    & \multicolumn{3}{c|}{\textbf{256 Tokens}}
    & \multicolumn{3}{c}{\textbf{512 Tokens}} \\
    \cmidrule(lr){2-4} \cmidrule(lr){5-7} \cmidrule(lr){8-10}
    \cmidrule(lr){11-13} \cmidrule(lr){14-16}
    & LPIPS $\downarrow$ & SSIM $\uparrow$ & PSNR $\uparrow$
    & LPIPS $\downarrow$ & SSIM $\uparrow$ & PSNR $\uparrow$
    & LPIPS $\downarrow$ & SSIM $\uparrow$ & PSNR $\uparrow$
    & LPIPS $\downarrow$ & SSIM $\uparrow$ & PSNR $\uparrow$
    & LPIPS $\downarrow$ & SSIM $\uparrow$ & PSNR $\uparrow$ \\
    \midrule

    OneDPiece
    & 0.372 & 0.379 & 15.66
    & 0.280 & 0.409 & 16.83
    & 0.212 & 0.450 & 18.09
    & 0.180 & 0.472 & 18.80
    & -- & -- & -- \\

    DOVE
    & 0.197 & 0.596 & 20.34
    & 0.168 & 0.619 & 21.06
    & 0.165 & 0.623 & 21.17
    & 0.153 & 0.633 & 21.54
    & -- & -- & -- \\

    ALIT
    & 0.301 & 0.350 & 15.86
    & 0.237 & 0.394 & 17.13
    & 0.186 & 0.443 & 18.42
    & 0.147 & 0.477 & 19.52
    & -- & -- & -- \\

    KARL
    & 0.401 & 0.371 & 15.42
    & 0.335 & 0.413 & 16.66
    & 0.237 & 0.489 & 18.81
    & 0.154 & 0.568 & 21.08
    & -- & -- & -- \\

    FlexTok
    & 0.434 & 0.327 & 14.24
    & 0.370 & 0.361 & 15.14
    & 0.290 & 0.443 & 17.16
    & 0.228 & 0.523 & 19.19
    & -- & -- & -- \\

    \rowcolor{gray!15}
    \textbf{Ours}
    & 0.344 & 0.363 & 15.72
    & 0.263 & 0.427 & 16.90
    & 0.201 & 0.497 & 18.17
    & 0.169 & 0.536 & 19.05
    & 0.153 & 0.556 & 19.56 \\

    \bottomrule
    \end{tabular}
    }
\end{table*}

\section{Performance Across Token Budgets}
Complementing our quantitative analysis in \cref{fig:main_quantitative_results},
\cref{tab:token_budget_results_1} presents additional reconstruction quality
metrics (LPIPS, SSIM, PSNR) across token budgets. Our method shows consistent perceptual quality improvement across token
budgets (LPIPS: $0.344 \rightarrow 0.153$, SSIM: $0.363 \rightarrow 0.556$).
At higher token counts (512), we achieve competitive fidelity with FlexTok
(LPIPS $0.153$ vs $0.228$) while maintaining $8.6\times$ faster decoding.

\section{Qualitative Analysis}
\begin{figure*}[t]
\centering
\includegraphics[width=\linewidth]{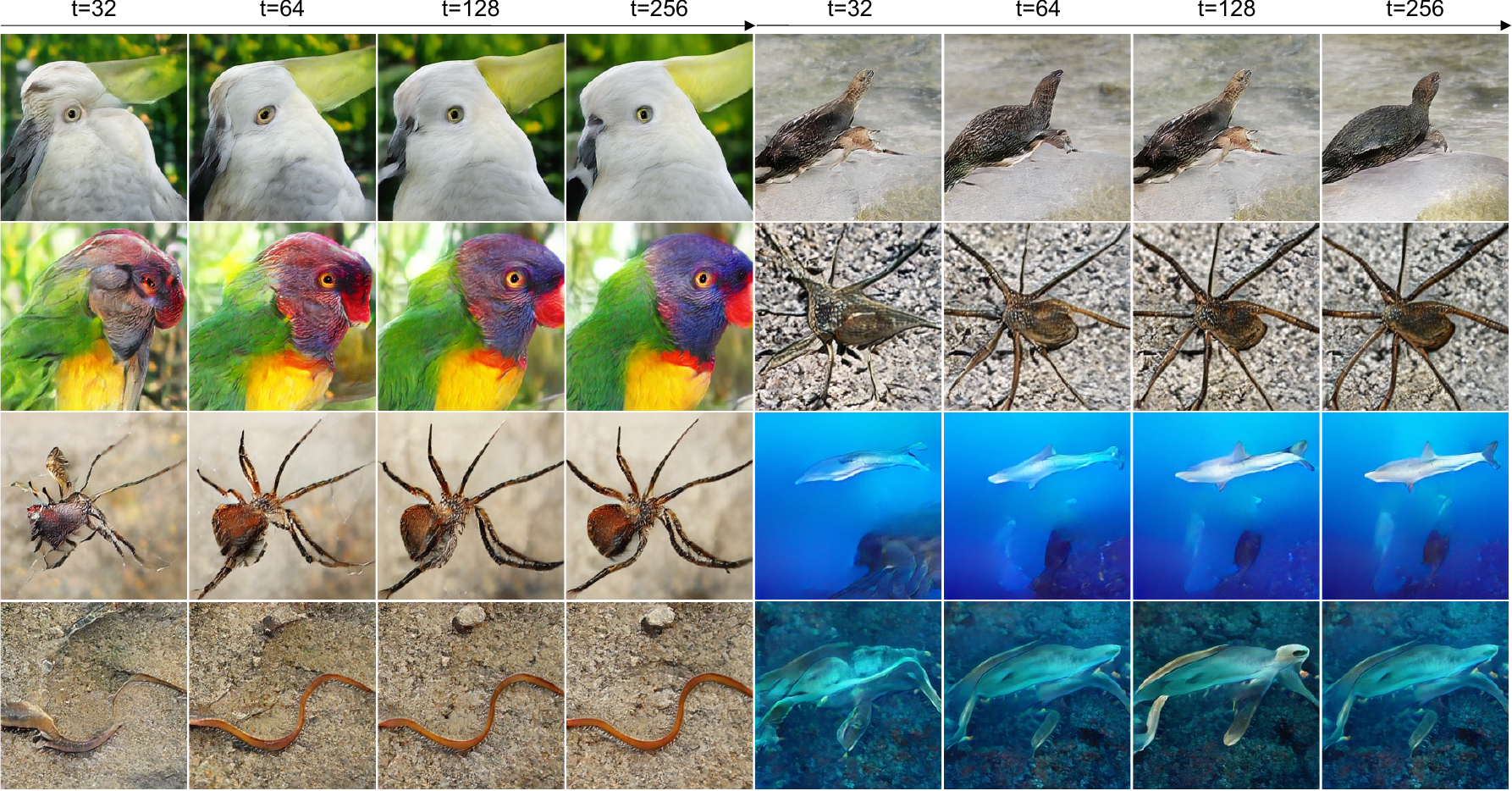}
\caption{\textbf{Autoregressive generation across token budgets.} LlamaGen~\cite{sun2024autoregressive} GPT-L generations across diverse ImageNet-100 categories (birds, insects, annelids, and marine life) using discrete channel tokens with truncated channels zero-filled. Even at 32 tokens, outputs maintain coherent global structure, with fidelity improving progressively at higher budgets. Generation at such low token counts is made possible by our tokenizer's channel ordering, which concentrates the most semantically meaningful information into the earliest tokens.}
\label{fig:ar_supp}
\end{figure*}

\begin{figure*}[tb]
\centering
\includegraphics[width=\linewidth]{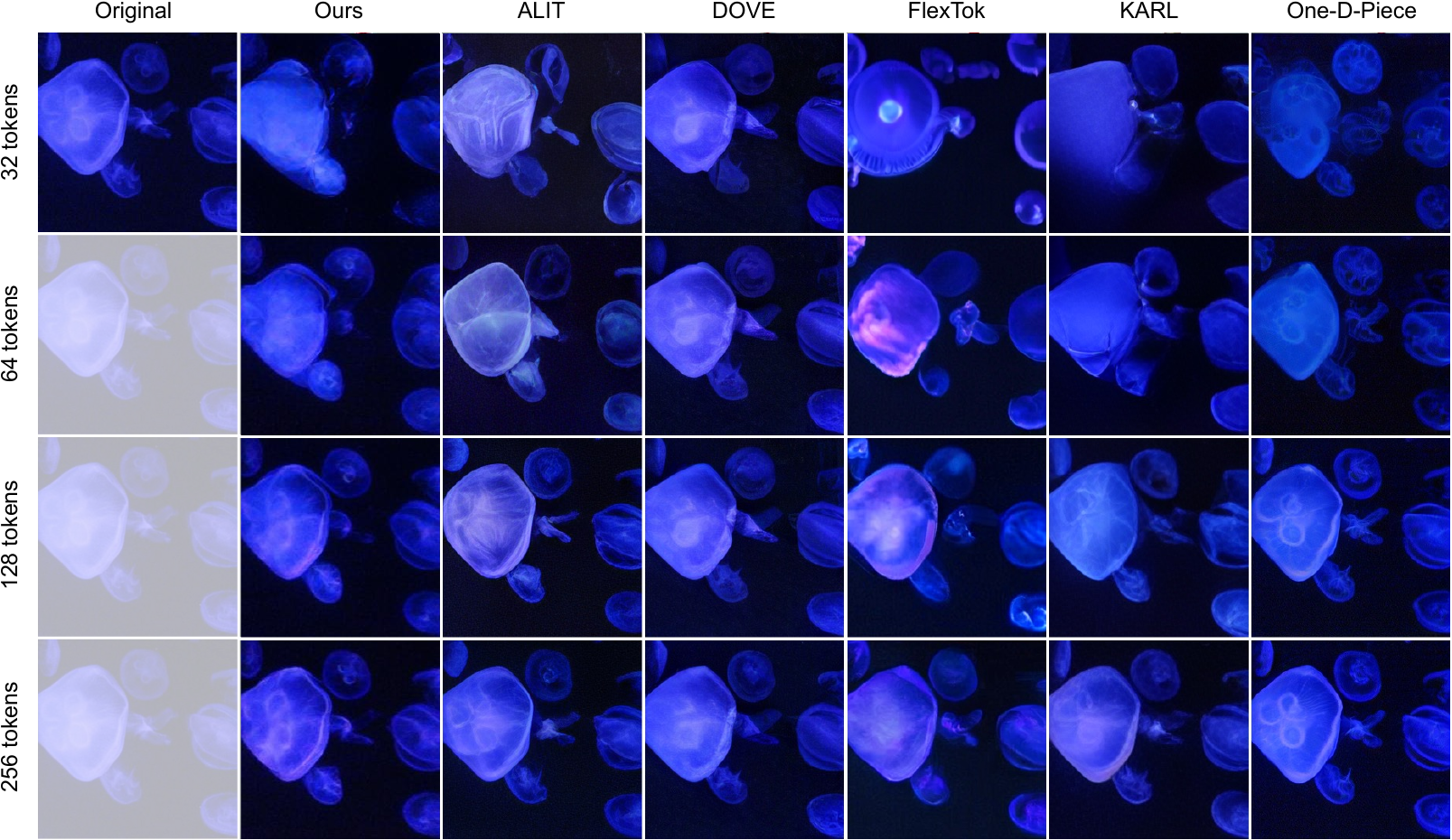}
\includegraphics[width=\linewidth]{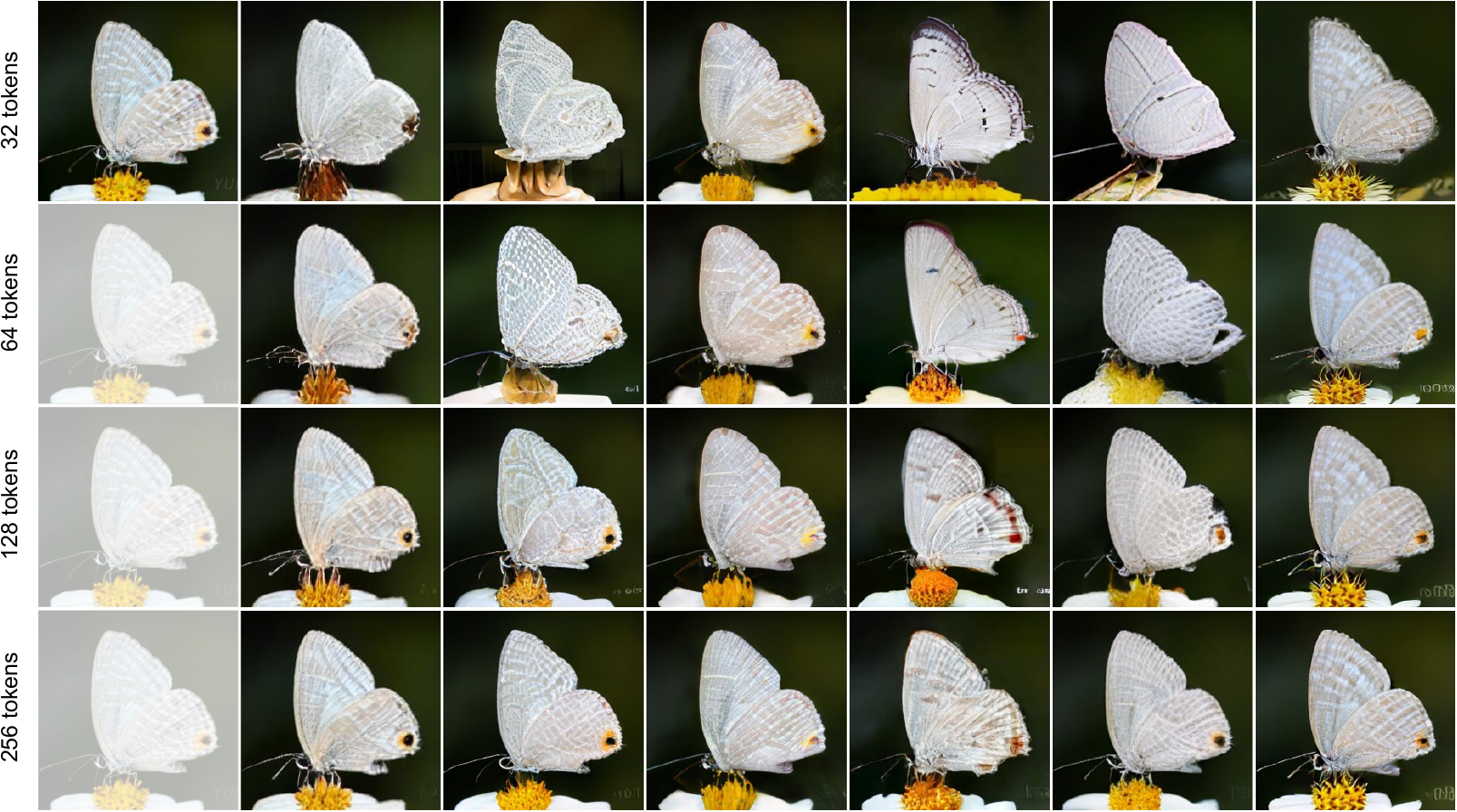}
\caption{\textbf{Qualitative comparison on images with contrasting tones.} Top: A jellyfish against a dark background, where our method preserves color fidelity even at lower token budgets. Bottom: A butterfly on a flower, where subtle wing textures and fine details emerge progressively with increasing tokens. Our method maintains perceptual coherence and colour consistency across all budgets.}
\label{fig:vertical_stack1}
\end{figure*}

\begin{figure*}[tb]
\centering
\includegraphics[width=\linewidth]{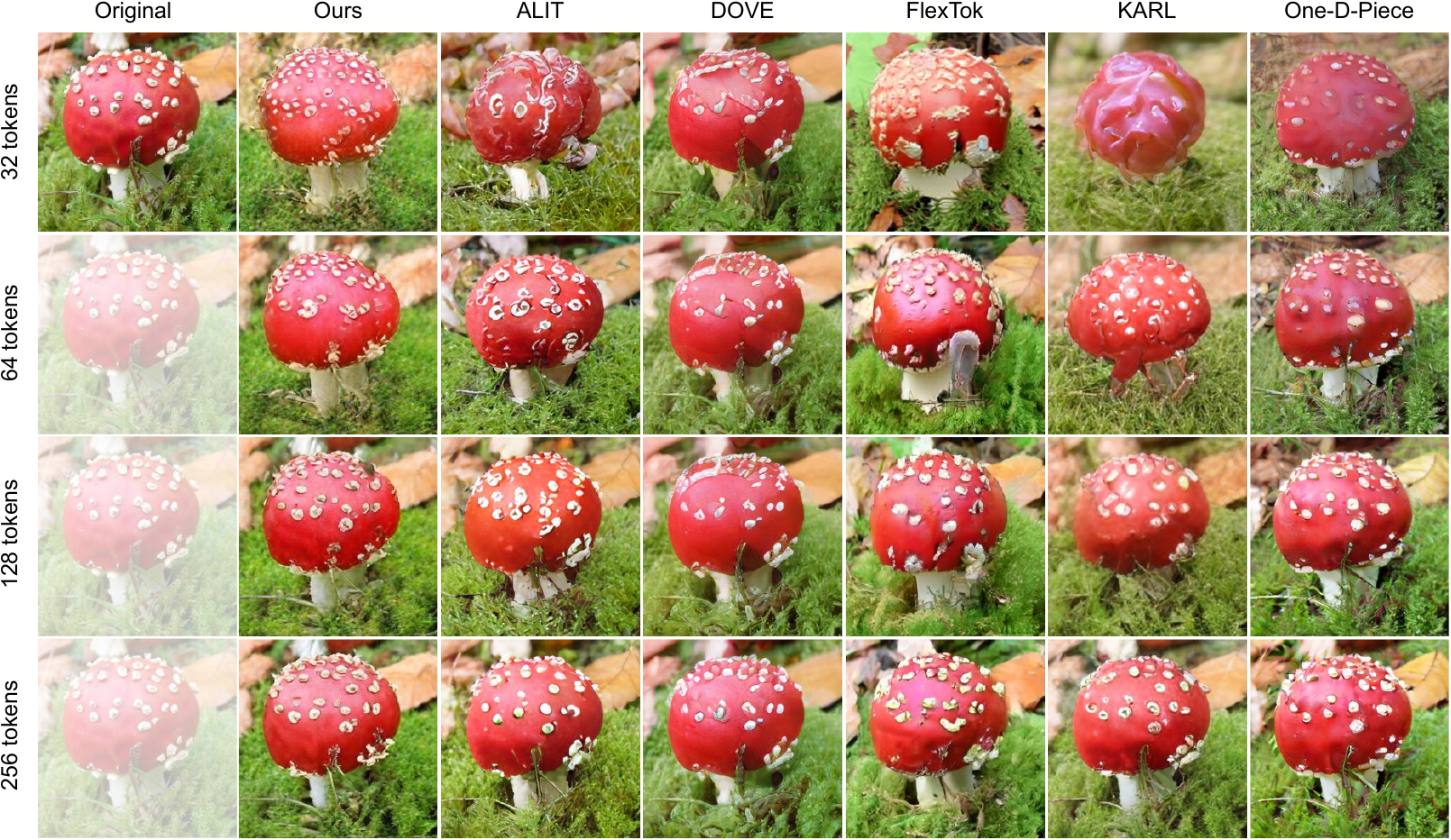}
\includegraphics[width=\linewidth]{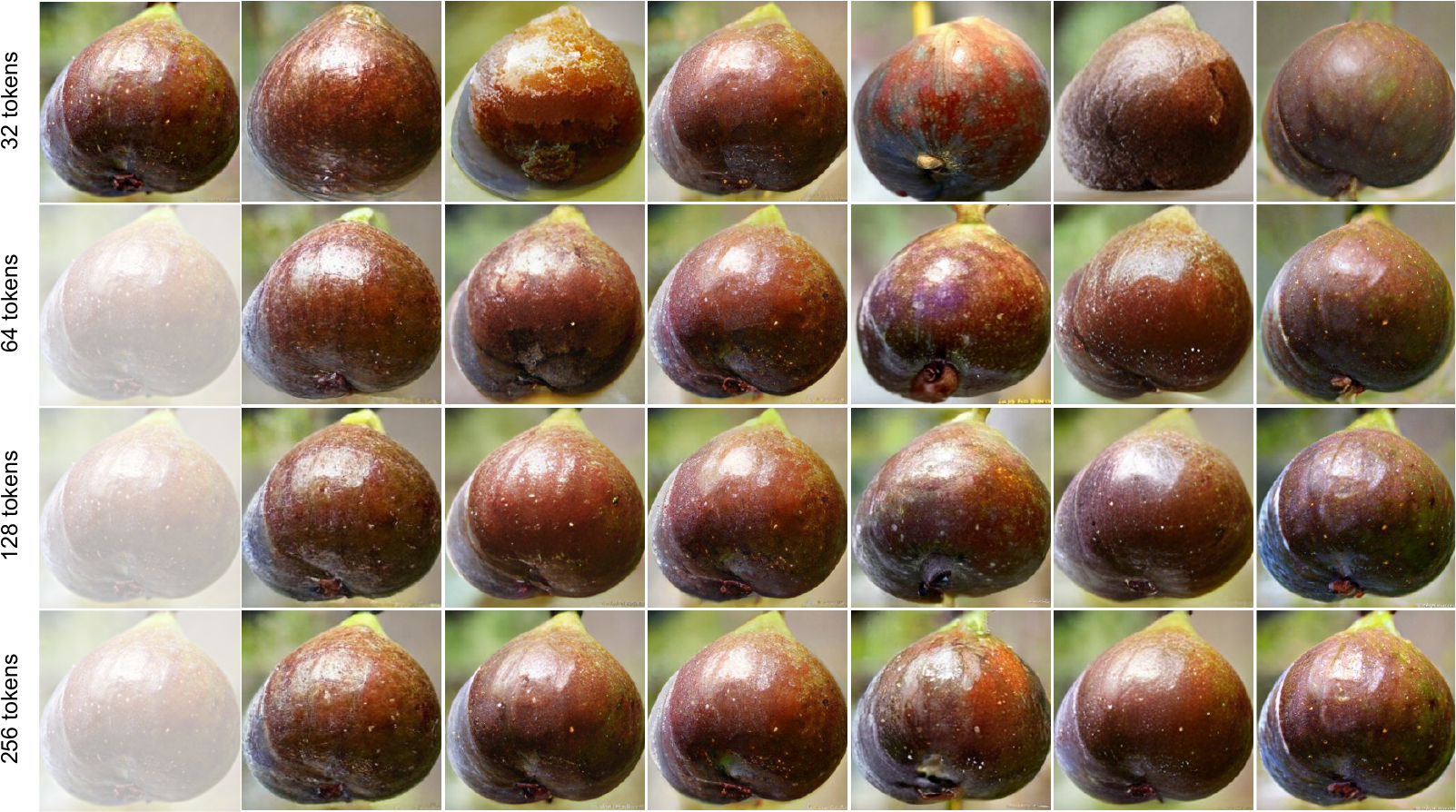}
\caption{\textbf{Qualitative comparison on images with varied textures.} Top: A red mushroom with white spots against a mossy background, where our method preserves fine surface detail and color fidelity even at low token budgets. Bottom: A dark round fruit, where competing methods introduce color artifacts and lose surface sheen at low tokens, while ours maintains perceptual consistency across all budgets.}
\label{fig:vertical_stack2}
\end{figure*}

\begin{figure*}[tb]
\centering
\includegraphics[width=\linewidth]{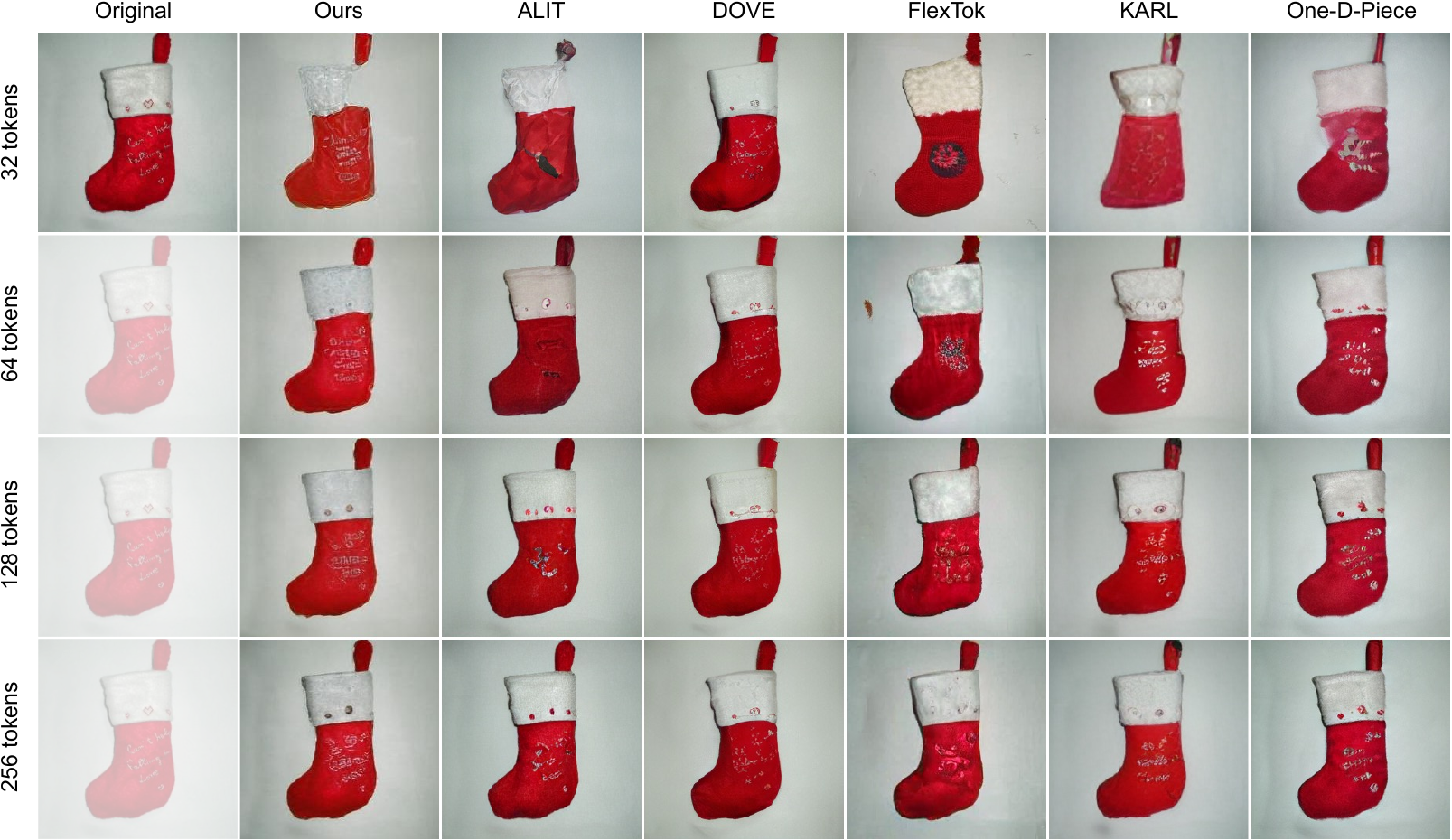}
\includegraphics[width=\linewidth]{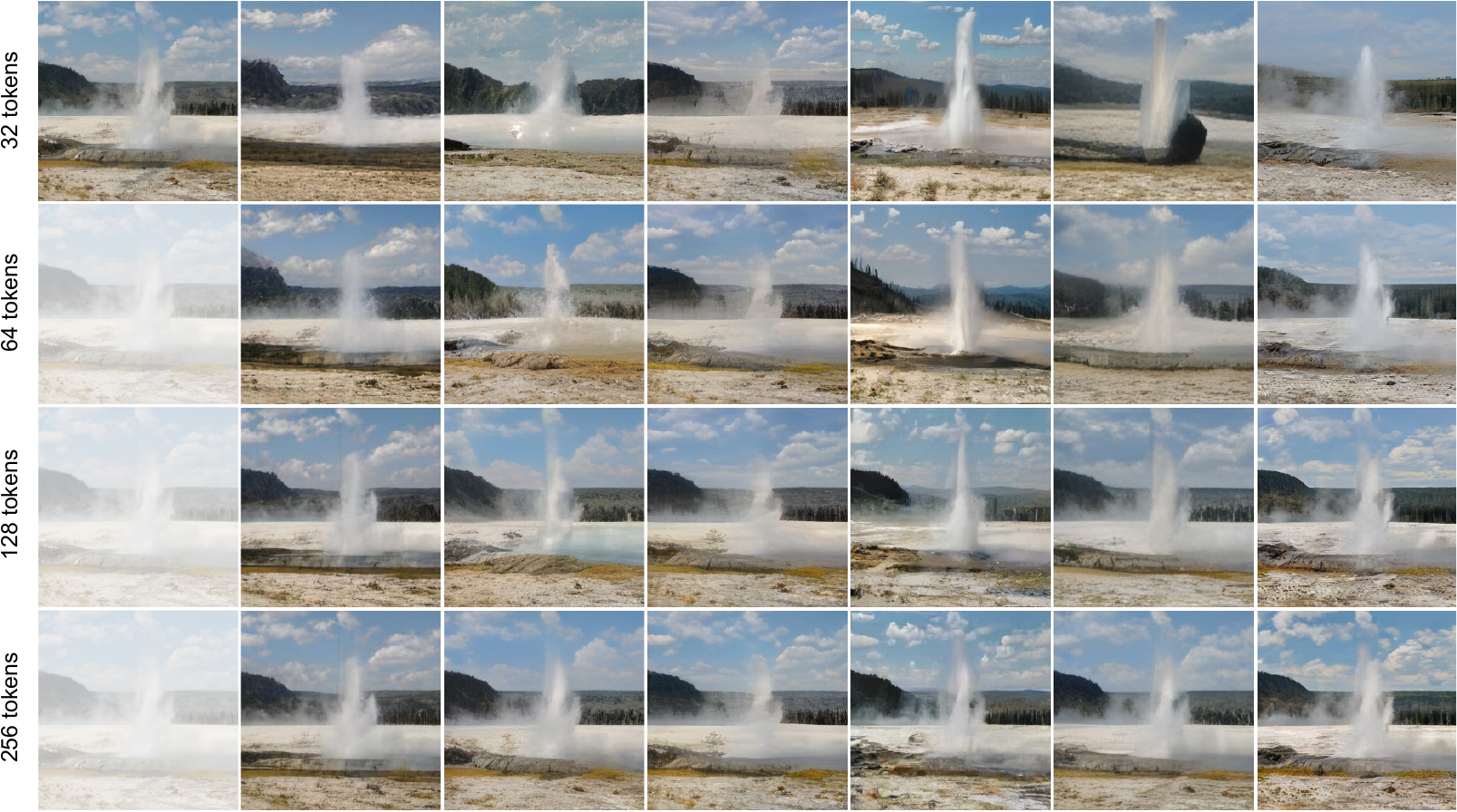}
\caption{\textbf{Reconstructions on cases with text and vibrant colours.} Top: Christmas stocking with text, where legibility remains difficult at low token counts but improves by 128 tokens. Bottom: A geyser eruption scene, where our method recovers landscape structure.}
\label{fig:vertical_stack3}
\end{figure*}

We evaluate reconstruction across diverse visual scenarios: contrasting foreground-background
(\cref{fig:vertical_stack1}), varied textures (\cref{fig:vertical_stack2}), scenes with vibrant colors and landscapes (\cref{fig:vertical_stack3}). Our method shows
consistent quality improvement with token budget, maintaining competitive fidelity
at 64--128 tokens. \cref{fig:channel_prog_supp} contrasts our flexible tokenizer
against a baseline without prefix channel masking. We also present qualitative autoregressive generation results across token budgets in~\cref{fig:ar_supp}, demonstrating coherent global structure even at 32 tokens.

\end{document}